\documentclass{article}

\usepackage[preprint,nonatbib]{neurips_2023}

\usepackage[utf8]{inputenc} 
\usepackage[T1]{fontenc}    
\usepackage{hyperref}       
\usepackage{url}            
\usepackage{booktabs}       
\usepackage{amsfonts}       
\usepackage{nicefrac}       
\usepackage{microtype}      
\usepackage{xcolor}         

\usepackage{graphicx}
\usepackage{amsmath,amsthm,amssymb}
\usepackage{color}
\usepackage[caption=false,farskip=0pt,labelfont={bf}]{subfig} 
\usepackage[small]{caption}
\usepackage{pifont}
\usepackage{diagbox}
\usepackage{multirow}
\usepackage{enumitem}
\usepackage{algorithm}
\usepackage{algorithmic}

\usepackage{mathtools}
\usepackage{makecell}
\usepackage{cancel}

\usepackage{amsmath,amsthm,amssymb}
\usepackage{color}

\newcommand{\beq}{\begin{equation}}
\newcommand{\eeq}{\end{equation}}
\newcommand{\beqs}{\begin{eqnarray}}
\newcommand{\eeqs}{\end{eqnarray}}
\newcommand{\barr}{\begin{array}}
	\newcommand{\earr}{\end{array}}
\newcommand{\bali}{\begin{aligned}}
	\newcommand{\eali}{\end{aligned}}

\newcommand{\Ac}[0]{\ensuremath{\mathcal{A}} }
\newcommand{\Bc}[0]{\ensuremath{\mathcal{B}} }

\newcommand{\Lc}[0]{\ensuremath{\mathcal{L}} }

\newcommand{\Nc}[0]{\ensuremath{\mathcal{N}} }

\newcommand{\Ebb}[0]{\ensuremath{\mathbb{E}} }

\newcommand{\Lbb}[0]{\ensuremath{\mathbb{L}} }

\newcommand{\Rbb}[0]{\ensuremath{\mathbb{R}} }
\newcommand{\Sbb}[0]{\ensuremath{\mathbb{S}} }
\newcommand{\Tbb}[0]{\ensuremath{\mathbb{T}} }

\newcommand{\Zbb}[0]{\ensuremath{\mathbb{Z}} }

\newcommand{\ie}[0]{\emph{i.e., }}

\newcommand{\eg}[0]{\emph{e.g., }}

\newcommand{\etc}[0]{\emph{etc. }}

\newcommand{\wrt}[0]{\emph{w.r.t. }}
\newcommand{\iid}[0]{\emph{i.i.d. }}

\newcommand{\Imat}[0]{\ensuremath{{\bf I}} }

\newcommand{\hv}[0]{\ensuremath{\boldsymbol{h}} }

\newcommand{\nv}[0]{\ensuremath{\boldsymbol{n}} }

\newcommand{\pv}[0]{\ensuremath{\boldsymbol{p}} }

\newcommand{\xv}[0]{\ensuremath{\boldsymbol{x}} }
\newcommand{\yv}[0]{\ensuremath{\boldsymbol{y}} }
\newcommand{\zv}[0]{\ensuremath{\boldsymbol{z}} }

\newcommand{\Av}[0]{\ensuremath{\boldsymbol{A}} }
\newcommand{\Bv}[0]{\ensuremath{\boldsymbol{B}} }

\newcommand{\Xv}[0]{\ensuremath{\boldsymbol{X}} }

\newcommand{\Omegamat}[0]{\ensuremath{\boldsymbol{\Omega}}}

\newcommand{\thetav}[0]{\ensuremath{\boldsymbol{\theta}} }

\newcommand{\muv}[0]{\ensuremath{\boldsymbol{\mu}} }

\newcommand{\phiv}[0]{\ensuremath{\boldsymbol{\phi}} }

\newcommand{\KL}[0]{\ensuremath{\mathrm{KL}} }
\newcommand{\JS}[0]{\ensuremath{\mathrm{JS}} }

\newcommand{\bb}[0]{\color{blue}}

\theoremstyle{plain}
\newtheorem{theorem}{Theorem}[section]

\theoremstyle{definition}

\theoremstyle{remark}
\newtheorem{remark}[theorem]{Remark}

\title{Big Learning}

\author{%
	Yulai Cong\thanks{Corresponding author: Yulai Cong <yulaicong@gmail.com>.} 
	\qquad\qquad
	Miaoyun Zhao
	\qquad
	\\
	{Sun Yat-sen University \qquad\quad  UBTECH \qquad\quad}
}

\begin{document}

\maketitle

\begin{abstract}

Recent advances in big/foundation models reveal a promising path for deep learning, where the roadmap steadily moves from big data to big models to (the newly-introduced) big learning.
Specifically, the big learning exhaustively exploits the information inherent in its large-scale \emph{complete/incomplete} training data, by simultaneously modeling many/all joint/conditional/marginal data distributions across potentially diverse domains, with one universal foundation model. 
We reveal that big learning ($i$) underlies most existing foundation models, ($ii$) is equipped with extraordinary flexibilities for complete/incomplete training data and trustworthy data tasks, ($iii$) is capable of delivering all joint/conditional/marginal data capabilities with one universal model, and ($iv$) unifies conventional machine learning paradigms and enables their flexible cooperations, manifested as a universal learning paradigm. 
Diverse experiments are carried out to validate the effectiveness of the presented big learning.

\end{abstract}

\section{Introduction}

AI is undergoing a paradigm shift with the rise of big/foundation models \cite{bommasani2021opportunities,yuan2022roadmap}, \eg BERT \cite{stickland2019bert}, GPT-3 \cite{brown2020language}, 
MAE \cite{he2021masked}, DALL-E \cite{ramesh2021zero,ramesh2022hierarchical}, Imagen \cite{saharia2022photorealistic}, Stable Diffusion \cite{rombach2022high}, UniDiffuser \cite{bao2023one}, ChatGPT \cite{ChatGPT,ouyang2022training}, \etc
Foundation models, often based on mask-and-predict pretraining and downstream finetuning, are capable of benefiting from pretraining on broad data at scale and accordingly, demonstrate diverse downstream task capabilities with impressive robustness \cite{stickland2019bert}, adaptability \cite{he2021masked}, and generalization \cite{ramesh2021zero}. 
Therefore, they are rapidly being integrated into real-world AI systems, \eg BERT into Google search,
Codex \cite{chen2021evaluating} into GitHub's Copilot, 
ChatGPT into Microsoft Bing, \etc

Despite the impressive capabilities and characteristics of foundation models, a unified theoretical framework justifying their great successes remains missing \cite{bommasani2021opportunities,yuan2022roadmap}, which is believed crucial for their further improvements and is likely a milestone for the foundation model community \cite{tamkin2021dabs}.
We propose the big learning to address that unified-framework challenge. 
In what follows, we first summarizing the reasons for the successes of existing foundation models and then we sublimate their training objectives into the presented big learning to benefit more from those reasons.

By referring to \cite{bommasani2021opportunities,yuan2022roadmap}, we attribute the successes of foundation models to  
the following two properties of their large-scale pretraining.
\begin{enumerate}[leftmargin=5mm]
    \item \textbf{Data comprehensiveness.} 
    Foundation models are often pretrained with massive data with great diversity.
    Often collected with minimal human interventions, these pretraining data are likely {comprehensively} consistent with the ``true'' data distribution that underlies both training/pretraining and test/finetuning phases,
    leading to a narrowed phase gap \emph{from the data perspective} and, therefore, serving as one reason for the generalization and robustness of foundation models.
    
    \item \textbf{Task comprehensiveness.} 
    Foundation models are pretrained in a massive multitasking manner on a wealth of \emph{data tasks}; \eg 
    both masked language modeling (MLM) and causal LM (CLM) leverage one universal model to simultaneously model many conditional data distributions (revealed in Section \ref{sec:our_method}).
    Such massive-task pretraining shows foundation models comprehensive task experience, which narrows the training-test/pretraining-finetuning gap \emph{from the task perspective} (it's likely the downstream task resembles a pretraining one).
    
\end{enumerate}

Inspired by existing foundation models succeeding from their comprehensive pretraining data and tasks, we propose the big learning that ($i$) is capable of enhancing both comprehensiveness to the extreme and ($ii$) unifies the training objectives of most foundation models within one framework.
Specifically, the big learning leverages a universal foundation model to simultaneously model \emph{many/all} joint/conditional/marginal data distributions across potentially diverse domains, manifested as a ``big'' \emph{generative}\footnote{
Throughout this paper, generative modeling is used in its broad sense; for example, classification may be viewed as the generative modeling of a label conditioned on its feature.
} learning task that exhaustively exploits the data information from many/all perspectives. 

The big learning naturally emerges from a promising path for deep learning, where the roadmap gradually follows \emph{big data, big models, big learning, $\cdots$}. 
Specifically, one collects \emph{big data} to comprehensively represent the underlying data distribution, develops \emph{big models} to serve as high-capacity information ``containers,'' relies on \emph{big learning} to comprehensively and exhaustively convey data information into those containers, and so forth.

Our contributions are summarized as follows.
\begin{itemize}[leftmargin=5mm]
    
    \item We propose the big learning to unify most foundation models within one framework.
    
    \item The big learning is capable of delivering \emph{all} joint/conditional/marginal data capabilities with one universal foundation model.
    Those data capabilities, in general settings, can manifest as classification, generation, completion/recommendation, \etc
    
    
    \item We empirically demonstrate that big learning ($i$) is feasible, ($ii$) delivers great model generalization, and ($iii$) can serve as a better strategy for finetuning foundation models.
    
\end{itemize}

\section{Preliminary}

\textbf{Big/Foundation models.} 
Taking shape in natural language processing (NLP), big/foundation models have drastically changed the research and practice of AI \cite{bommasani2021opportunities,yuan2022roadmap}. 
BERT \cite{stickland2019bert} and GPT series \cite{radford2019language,brown2020language} significantly accelerate the development of NLP, while models like DALL-Es \cite{ramesh2021zero,ramesh2022hierarchical}, Stable Diffusion \cite{rombach2022high}, and UniDiffuser \cite{bao2023one} effectively promote interdisciplinary research among different research fields, initiating a new revolution of AI-Generated Content (AIGC).

Most existing foundation models are pretrained\footnote{
See Appendix \ref{appsec:biglearn_contrastlearn} for discussions on pretraining associated with contrastive learning.
} with ($i$) masked LM (or masked auto-encoding; like BERT and MAE), ($ii$) causal/auto-regressive LM (like GPTs and DALL-E), and ($iii$) permutation LM (like XLNET \cite{yang2019xlnet}).
We will demonstrate in Section \ref{sec:our_method} that these pretraining methods are all special cases of the proposed big learning, which, accordingly, serves as a unified theoretical framework that reveals one underlying principle of foundation models.

\textbf{Transformers and Vision Transformers (ViTs).} 
Based on the flexible self-attention mechanism \cite{vaswani2017attention}, Transformers have been serving as the de facto model architecture for foundation models.
Often Transformers take as input an $L$-length sequence of discrete tokens $\xv\in \Zbb^{L}$ and output the corresponding $D$-dimensional embedding $\hv \in \Rbb^{L\times D}$, with the self-attention mechanism flexibly customized (among the $L$ locations) to implement masked/causal/permutation LM.
ViTs \cite{dosovitskiy2020image} are Transformers modified for modeling continuous image patches. 
Despite their high model capacity and flexible modeling capabilities, Transformers/ViTs are well-known to be over-parameterized and data/information hungry \cite{lan2019albert,hassani2021escaping,wang2022towards}; 
we will reveal that those properties of Transformers/ViTs exactly matches the big learning.

\textbf{Multi-mode training objectives.} 
Two well-known multi-mode training objectives are ($i$) the cross-entropy loss, often used in maximum likelihood learning with \emph{discrete} categorical observations, and ($ii$) the GAN loss \cite{goodfellow2014generative} for adversarial learning on \emph{continuous} observations, as detailed below.
\begin{enumerate}[leftmargin=5mm]
	
	\item \textbf{The cross-entropy loss.} Given history observations $\xv$ and the current word $y$ sampled from the underlying data distribution $q(\xv,y)$, and a model $p_{\thetav}(y|\xv)$ modeling the categorical distribution of $y$ given $\xv$, the commonly-used cross-entropy loss is identical to 
	\beq\label{eq:cross_entropy_loss}
	\bali
		\Ebb_{q(\xv,y)} [- \log p_{\thetav}(y|\xv)]
		& \propto \KL[q(\xv,y)||p_{\thetav}(y|\xv)q(\xv)],
	\eali
	\eeq
	where the optimal $p_{\thetav^{*}}(y|\xv)=q(y|\xv)$. 
	Note the categorical modeling of $p_{\thetav}(y|\xv)$ can model multiple modes\footnote{
	A misunderstanding is that $p_{\thetav}(y|\xv)$ has to be uni-model under the classification setup with feature $\xv$ and label $y$. Note a multi-mode model can have a uni-model practical instantiation.
	}, \eg consider the diverse generation from the GPT-$3$ \cite{brown2020language}.
	
	\item \textbf{The GAN loss.} GANs are known for synthesizing highly realistic images with multiple modes \cite{Karras_2019_CVPR,karras2020analyzing,karras2021alias}.
	A standard GAN consists of a generator $G_{\thetav}$ and a discriminator $D_{\phiv}$, both of which are trained in an adversarial manner via
	\beq\label{eq:standard_GAN_loss}
	\min_{\thetav} \max_{\phiv} \Ebb_{q(\xv)} \log D_{\phiv}(\xv) + \Ebb_{p_{\thetav}(\xv)} \log (1-D_{\phiv}(\xv)) ,
	\eeq
	where $q(\xv)$ is the underlying data distribution and $p_{\thetav}(\xv)$ is the generated distribution with the generative process $\xv=G_{\thetav}(\zv), \zv\sim p(\zv)$. $p(\zv)$ is an easy-to-sample distribution, like a normal distribution. 
	With optimal $D_{\phiv^{*}}$, Eq. \eqref{eq:standard_GAN_loss} minimizes the Jensen-Shannon (JS) divergence $\JS[q(\xv)||p_{\thetav}(\xv)]$ \cite{goodfellow2014generative}.
	
\end{enumerate}

To demonstrate the flexibilities of the big learning, we instantiate it within both maximum likelihood and adversarial learning territories (with the multi-mode objectives in \eqref{eq:cross_entropy_loss} and \eqref{eq:standard_GAN_loss}, respectively) in Section \ref{sec:example_unsupervised_biglearn}, where Transformers/ViTs are employed to construct its universal foundation model.

\section{Big Learning}
\label{sec:our_method}

For better introduction of the big learning, we first present its main idea in simplified unsupervised/uni-modal settings, where a data sample $\Xv = (\xv)$ contains only a feature $\xv\in \Rbb^{L\times D}$ (with length $L$ and dimension $D$; like $L$ words with $D=1$ or $L$ flattened image patches with $D$ pixels); 
we then generalize its scope to the general settings, where a data sample $\Xv = (\yv, \xv)$ contains both feature $\xv$ and its paired supervision $\yv \in \Rbb^{L^{\yv}\times D^{\yv}}$ (\eg when $L^{\yv}=D^{\yv}=1$, $y \in \{1, \cdots, C\}$ may represent a label). 
In both settings, the big learning can naturally handle ``incomplete data,'' which are defined as either $\xv$ missing values along the $L$-dimension or $\yv$ missing values along the $L^{\yv}$-dimension.

\subsection{Unsupervised/Uni-modal Big Learning}
\label{sec:unsupervised_biglearn}

Given complete data samples $\xv\in \Rbb^{L\times D}$ drawn from the underlying data distribution $q(\xv)$, the mainstream machine learning paradigms concentrate on \emph{joint modeling}, \ie to construct a model $p_{\thetav}(\xv)$ in the joint domain to resemble $q(\xv)$, or informally $p_{\thetav}(\xv) \longrightarrow q(\xv)$, leveraging objectives like GANs \cite{brock2018large,Karras_2019_CVPR}, VAEs \cite{kingma2013auto,dai2018diagnosing}, Flows \cite{dinh2014nice,kingma2018glow}, diffusion models \cite{ho2020denoising,song2020score}, \etc

However, \emph{joint modeling} can not take advantage of incomplete data (\eg $\xv$ missing values along the $L$-dimension), which frequently arise in practical applications. Moreover, it may also fail to comprehensively exploit the information from a complete data sample, because the corresponding conditional/marginal samples (already given within that joint sample) are not explicitly utilized.

Intuitively, one would prefer to use the available complete/incomplete data as they are (\eg to avoid unintentional interventions that violate the \iid assumption) and to comprehensively exploit the data information therein (\eg to from multiple training tasks that encourage learning compositional intrinsic data knowledge in model parameters \cite{lu2021pretrained,aghajanyan2021muppet}).
The latter has been extensively demonstrated by existing foundation models, which exploit data information from many \emph{conditional} perspectives in a multitasking manner and demonstrate impressive robustness and generalization. 
For clarity, Table \ref{tab:BigLearn_special_cases} summarizes the commonly-used training objectives for foundation models, \ie masked language modeling (LM), causal/auto-regressive LM, and permutation LM.

\begin{figure}
	\vspace{-5mm}
	\centering
	\subfloat[]{
		\includegraphics[height=0.32\columnwidth]{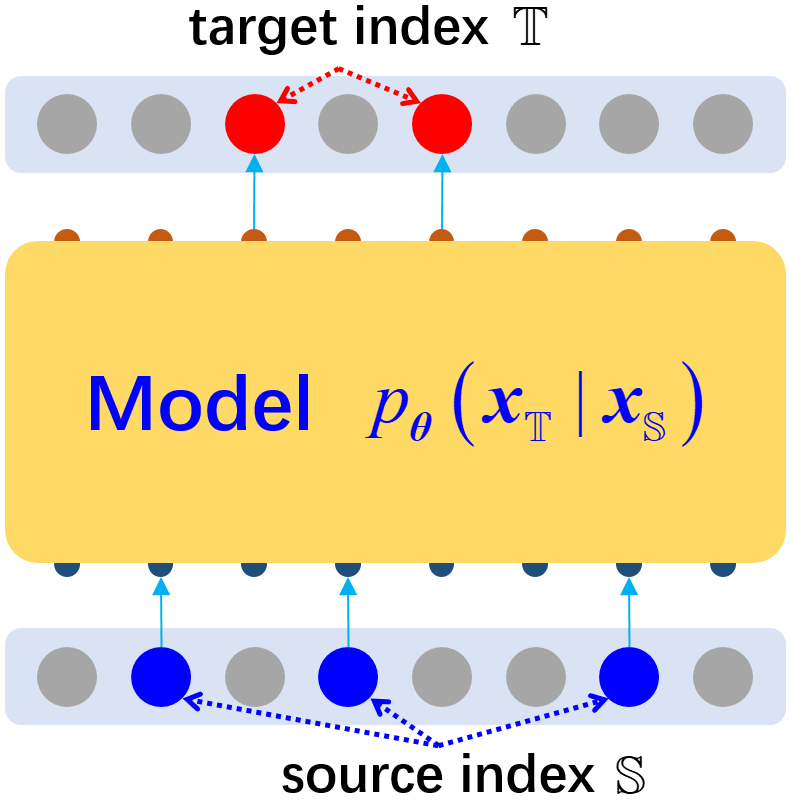}
		\label{fig:arch_unsupervised_biglearning}}
	\qquad\qquad\qquad
	\subfloat[]{
		\includegraphics[height=0.32\columnwidth]{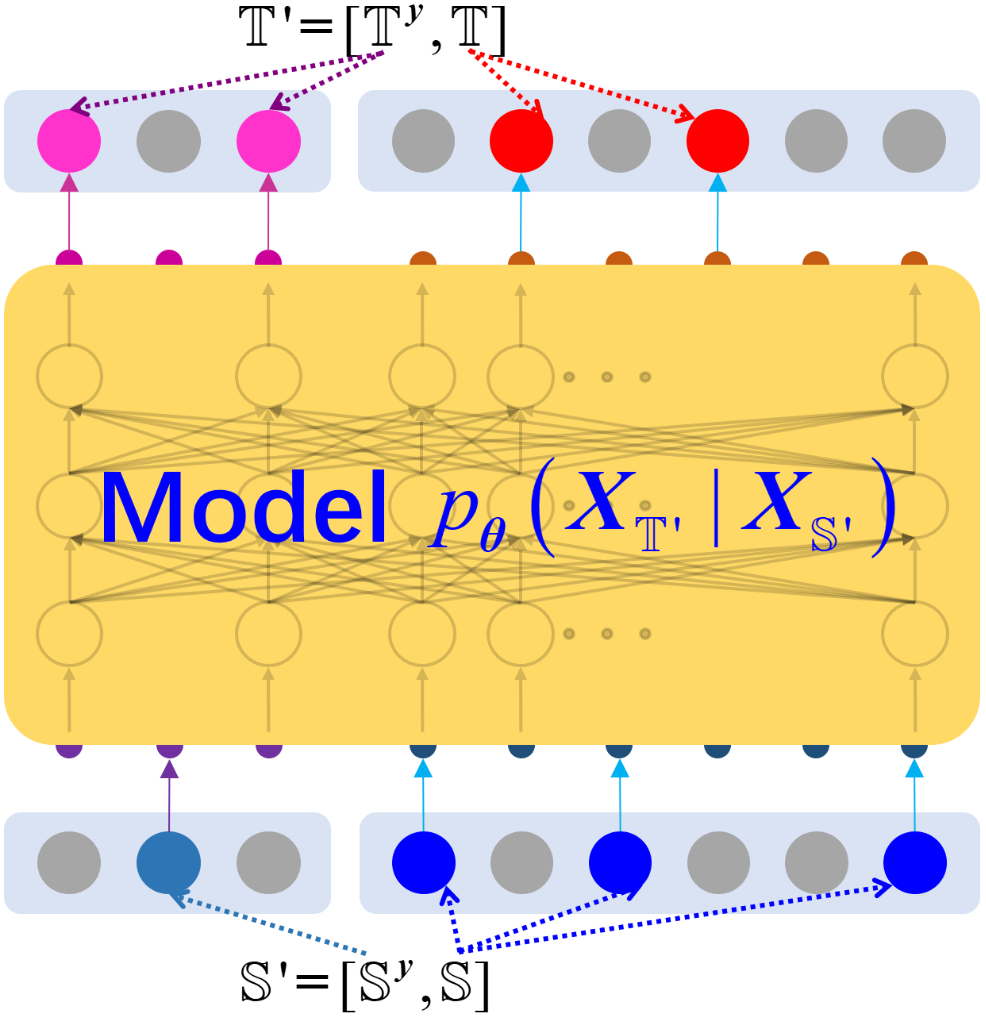}
		\label{fig:arch_big_learning}}
	\caption{Demonstration of the unsupervised/uni-modal big learning (a) and the big learning (b).}
	\vspace{-2mm}
\end{figure}

To comprehensively exploit the data information within both complete and incomplete samples, we propose the following unsupervised big learning that leverages a universal foundation model to model many/all joint/conditional/marginal data distributions\footnote{
The incomplete data are readily utilized in the corresponding conditional/marginal tasks.
} simultaneously (manifested as \emph{``big'' learning with massive data tasks}).

\begin{theorem}[Unsupervised/Uni-modal big learning]
	\label{theom:unsupervised_biglearn}
	With the unsupervised/uni-modal setup, where a data sample $\Xv = (\xv)$ contains only a feature $\xv\in \Rbb^{L\times D}$ with length $L$, dimension $D$, the length index set $\Lbb=\{1,\cdots,L\}$, and any two {non-overlapping} subsets
	$\Sbb \subset \Lbb$ and $\Tbb \subseteq \Lbb, \Tbb \neq \emptyset$, unsupervised big learning leverages a universal foundation model $p_{\thetav}(\xv_{\Tbb}|\xv_{\Sbb})$ (see Fig. \ref{fig:arch_unsupervised_biglearning}) to model many/all joint/conditional/marginal data distributions\footnote{
		After training, the universal model $p_{\thetav}(\xv_{\Tbb}|\xv_{\Sbb})$ is expected to possess many/all joint/conditional/marginal data capabilities. 
			Note to naively collect all the capabilities, one need construct at least $N_{\text{all}}=\sum_{i=0}^{L-1} C_L^{i} (\sum_{k=1}^{L-i} C_{L-i}^k)$ models, which is clearly prohibitive. See Appendix \ref{appsec:naive_model_big_learning} for details.
	} simultaneously, \ie 
	\beq\bali\label{eq:uni_model}
		& p_{\thetav}(\xv_{\Tbb}|\xv_{\Sbb})  \longrightarrow q(\xv_{\Tbb}|\xv_{\Sbb}), \forall (\Sbb,\Tbb) \in \Omegamat
	\eali\eeq
	where $\Sbb$ and $\Tbb$ are named source/target index subsets, respectively, and $\Omegamat$ can be flexibly specified as the set that contains all $(\Sbb,\Tbb)$ pairs or a portion of them.
	For any $(\Sbb,\Tbb)$, $q(\xv_{\Tbb}|\xv_{\Sbb})$ denotes the corresponding joint/conditional/marginal data distribution, whose samples are readily selected from the whole available data.
\end{theorem}

\begin{remark}
	In Theorem \ref{theom:unsupervised_biglearn}, $\Sbb \cup \Tbb$ need not be $\Lbb$, meaning that incomplete data are naturally utilized.
\end{remark}

\begin{remark}
	\vspace{1mm}
	Because source $\xv_{\Sbb}$ and target $\xv_{\Tbb}$ may have different sizes for different $(\Sbb,\Tbb) \in \Omegamat$, one may prefer constructing $p_{\thetav}(\xv_{\Tbb}|\xv_{\Sbb})$ of \eqref{eq:uni_model} as a Transformer/ViT to embrace its modeling flexibility.
	\vspace{1mm}
\end{remark}

\begin{remark}
	The objective that measures the distance/divergence between both sides of \eqref{eq:uni_model} is application-dependent; possible choices include the cross-entropy loss, the GAN loss, energy-based models \cite{lecun2022path}, diffusion models, score matching \cite{hyvarinen2005estimation}, \emph{etc}.
	\vspace{1mm}
\end{remark}

\begin{remark}
Considering practical situations, one may alternatively or additionally do big learning in transformed domains, \eg via 
$p_{\thetav}(\hat \xv_{\Tbb}|\hat \xv_{\Sbb})$ with $\hat\xv=g(\xv)$ 
or 
$p_{\thetav}(h(\xv_{\Tbb})|k(\xv_{\Sbb}))$ \cite{he2021masked,wei2021masked},
where $g(\cdot)$, $h(\cdot)$, and $k(\cdot)$ are domain-knowledge-inspired transformations.
\end{remark}

\subsection{Implementations of Unsupervised/Uni-modal Big Learning}
\label{sec:example_unsupervised_biglearn}

We demonstrate unsupervised/uni-modal big learning with two example implementations, one of which is in the adversarial-learning territory with continuous observations, while the other is in the maximum-likelihood-learning territory with discrete observations.

\begin{table}
	\centering
	\caption{Big learning and its special cases. In general, $\Xv = (\yv, \xv)$, $\xv \in \Rbb^{L\times D}$, $\yv \in \Rbb^{L^{\yv}\times D^{\yv}}$, $\Lbb' = [\Lbb^{\yv},\Lbb]$, $\Sbb' = [\Sbb^{\yv}, \Sbb]$, and $\Tbb' = [\Tbb^{\yv}, \Tbb]$; with $\Xv = (\xv)$ and $\Lbb^{\yv}=\Sbb^{\yv}=\Tbb^{\yv}=\emptyset$, unsupervised big learning is recovered.
		When $y \in \{1, \cdots, C\}^{1\times 1}$, it may represent a label.
		We ignore the implementation details and only focus on the core idea for demonstration.
	}
	\resizebox{\columnwidth}{!}{
		\begin{tabular}{l c l}
			\hline\hline 
			\multirow{2}{*}{Big Learning} & \multirow{2}{*}{
				$p_{\thetav}(\Xv_{\Tbb'}|\Xv_{\Sbb'}) \longrightarrow q(\Xv_{\Tbb'}|\Xv_{\Sbb'}), \forall ({\Sbb'},{\Tbb'})$
			} & 
			\multirow{2}{*}{$\Sbb' \subset \Lbb', \Tbb' \subseteq \Lbb', \Tbb' \neq \emptyset,$ and $\Sbb'\cap\Tbb' = \emptyset$ } \\
			& & \\
			\hline\hline 
			$\qquad\downarrow$Special Case & $\downarrow$Training Objective & $\qquad\downarrow$Constraints \\
			\hline\hline			
			\multirow{3}{*}{\makecell[l]{Masked LM\\{\cite{stickland2019bert}}}} & \multirow{3}{*}{
				$\Ebb_{q(\Sbb,\Tbb)} \KL[q(\xv_{\Tbb}|\xv_{\Sbb})||p_{\thetav}(\xv_{\Tbb}|\xv_{\Sbb})]$
			} & 
			\multirow{3}{*}{\makecell[l]{
					$q(\Sbb,\Tbb) = \mathcal{U}\{(\Sbb,\Tbb): \Sbb$ is a $85\%$ random subset of $\Lbb$, and $\Tbb=\Lbb\backslash\Sbb\}$
					\\
					$p_{\thetav}(\xv_{\Tbb}|\xv_{\Sbb}) = \prod_{t\in\Tbb} \text{Categorical}(x_{t}|\pv_{\thetav}(\xv_{\Sbb}))$
			}} \\
			& & \\
			& & \\
			\hline
			\multirow{3}{*}{\makecell[l]{Causal/Auto-regressive LM\\\cite{brown2020language,ramesh2021zero,ChatGPT,ouyang2022training}}} & \multirow{3}{*}{
				$\sum_{(\Sbb,\Tbb) \in \Xi} \KL[q(\xv_{\Tbb}|\xv_{\Sbb}) || p_{\thetav}(\xv_{\Tbb}|\xv_{\Sbb})]$
			} & 
			\multirow{3}{*}{\makecell[l]{
					$\Xi = \{(\emptyset, 1), (\{1\}, 2), (\{1,2\}, 3), \cdots\}$ 
					\\
					$p_{\thetav}(\xv_{\Tbb}|\xv_{\Sbb}) = \text{Categorical}(\xv_{\Tbb}|\pv_{\thetav}(\xv_{\Sbb}))$
			}} \\
			& & \\
			& & \\			
			\hline			
			\multirow{5}{*}{\makecell[l]{Permutation LM\\\cite{yang2019xlnet}}} & 
			\multirow{5}{*}{
				$\Ebb_{q(\Sbb,\Tbb)} \sum_{(\bar\Sbb,\bar\Tbb) \in \Xi_{\Sbb,\Tbb}} \KL[q(\xv_{\bar\Tbb}|\xv_{\bar\Sbb}) || p_{\thetav}(\xv_{\bar\Tbb}|\xv_{\bar\Sbb})]$
			} & 
			\multirow{5}{*}{\makecell[l]{
					$q(\Sbb,\Tbb) = \mathcal{U}\{(\Sbb,\Tbb): \Sbb$ is a $85\%$ random subset of $\Lbb$, and 
					\\
					\qquad\qquad\qquad\qquad $\Tbb=\{t_1,t_2,\cdots\}$ is a random permutation of  $\Lbb\backslash\Sbb\}$
					\\
					$\Xi_{\Sbb,\Tbb} = \{(\Sbb,t_1), (\{\Sbb,t_1\},t_2), (\{\Sbb,t_1, t_2\},t_3), \cdots\}$
					\\ $p_{\thetav}(\xv_{\bar\Tbb}|\xv_{\bar\Sbb}) =\text{Categorical}(\xv_{\bar\Tbb}|\pv_{\thetav}(\xv_{\bar\Sbb}))$
			}} \\
			& & \\
			& & \\
			& & \\
			& & \\
			\hline
			\multirow{3}{*}{\makecell[l]{MAE \cite{he2021masked}\\MaskFeat \cite{wei2021masked}}} & 
			\multirow{3}{*}{
				$\Ebb_{q(\Sbb, \Tbb)} \KL[q(h(\xv_{\Tbb})|\xv_{\Sbb})||p_{\thetav}(h(\xv_{\Tbb})|\xv_{\Sbb})]$
			} & 
			\multirow{3}{*}{\makecell[l]{
					$q(\Sbb, \Tbb) = \mathcal{U}\{(\Sbb, \Tbb): \Sbb$ is a $25\%$ random subset of $\Lbb$, and $\Tbb = \Lbb\backslash\Sbb$
					\\
					$p_{\thetav}(h(\xv_{\Tbb})|\xv_{\Sbb}) = \Nc(h(\xv_{\Tbb})|\muv_{\thetav}(\xv_{\Sbb}), \Imat)$
					\\
					$h(\cdot)$ is a normalization/HOG transformation for MAE/MaskFeat
			}} \\
			& & \\
			& & \\
			\hline	
			\multirow{4}{*}{\makecell[l]{
					Big Learning
					with \eqref{eq:model_to_data_all}
			}} & 		
			\multirow{4}{*}{
				$\Ebb_{q(\Sbb, \Tbb)} \JS[q(\xv_{\Sbb\cup\Tbb})||p_{\thetav}(\xv_{\Tbb} | \xv_{\Sbb}) q(\xv_{\Sbb})]$
			} & 
			\multirow{4}{*}{\makecell[l]{
					$q(\Sbb,\Tbb) = \mathcal{U}\{(\Sbb,\Tbb): \Sbb$ is a random subset of $\Lbb$, and $\Tbb$ is a random 
					\\
					\qquad\qquad\qquad\qquad\quad subset of $\Lbb\backslash\Sbb\}$
					\\
					$p_{\thetav}(\xv_{\Tbb}|\xv_{\Sbb})$ is a universal ViT-based GAN generator
			}} \\
			& & \\
			& & \\
			& & \\
			\hline
			\multirow{5}{*}{\makecell[l]{
					Big Learning
					with \eqref{eq:ML_implementation}
			}} & 
			\multirow{5}{*}{
				$\Ebb_{q(\Sbb,\Tbb)} \sum_{(\bar\Sbb,\bar\Tbb) \in \Xi_{\Sbb,\Tbb}} \KL[q(\xv_{\bar\Tbb}|\xv_{\bar\Sbb}) || p_{\thetav}(\xv_{\bar\Tbb}|\xv_{\bar\Sbb})]$
			} & 
			\multirow{5}{*}{\makecell[l]{
					$q(\Sbb,\Tbb) = \mathcal{U}\{(\Sbb,\Tbb): \Sbb$ is a random subset of $\Lbb$, and 
					\\
					\qquad\qquad\qquad $\Tbb=\{t_1,t_2,\cdots\}$ is a random permuted subset of $\Lbb\backslash\Sbb\}$
					\\
					$\Xi_{\Sbb,\Tbb} = \{(\Sbb,t_1), (\{\Sbb,t_1\},t_2), (\{\Sbb,t_1, t_2\},t_3), \cdots\}$
					\\ $p_{\thetav}(\xv_{\bar\Tbb}|\xv_{\bar\Sbb}) =\text{Categorical}(\xv_{\bar\Tbb}|\pv_{\thetav}(\xv_{\bar\Sbb}))$
			}} \\
			& & \\
			& & \\
			& & \\
			& & \\
			\hline \hline
			Supervised Classification & \eg $\KL[q(\yv|\xv)||p_{\thetav}(\yv|\xv)]$ & $\Sbb' = [\emptyset, \Lbb], \Tbb' = [\Lbb^{\yv}, \emptyset], p_{\thetav}(\yv|\xv)$ is \eg a classifier \\
			Joint Generation & \eg $\JS[q(\xv)||p_{\thetav}(\xv)]$ & $\Sbb' = [\emptyset, \emptyset], \Tbb' = [\emptyset, \Lbb]$, $p_{\thetav}(\xv)$ may be a generator \\
			Conditioned Generation & \eg $\KL[q(\xv|\yv)||p_{\thetav}(\xv|\yv)]$ & $\Sbb' = [\Lbb^{\yv}, \emptyset], \Tbb' = [\emptyset, \Lbb]$, $p_{\thetav}(\xv|\yv)$: a conditional flow\\
			\hline \hline
		\end{tabular}
	}
	\label{tab:BigLearn_special_cases}
		\vspace{-2mm}
\end{table}

%

\subsubsection{Adversarial Learning for Foundation Models}
\label{sec:GAN_implementation}

Given continuous observations $\xv \in \Rbb^{L\times D}$ (\eg $\xv$ denoting a sequence of flattened patches of an image), we design a ViT-based universal model $p_{\thetav}(\xv_{\Tbb}|\xv_{\Sbb})$ that models the generative processes of $\xv_{\Tbb}$ given $\xv_{\Sbb}$ for \emph{all} $(\Sbb,\Tbb)$ pairs (see Appendix \ref{appsec:GAN_arch} for the detailed architecture).
The standard GAN loss \cite{goodfellow2014generative} is employed as the objective that measures the divergence between both sides of \eqref{eq:uni_model}.

Following \eqref{eq:uni_model} of Theorem \ref{theom:unsupervised_biglearn}, we propose to adversarially train the foundation model $p_{\thetav}(\xv_{\Tbb}|\xv_{\Sbb})$ via minimizing the JS divergence between $p_{\thetav}(\xv_{\Tbb}|\xv_{\Sbb})$ and $q(\xv_{\Tbb}|\xv_{\Sbb})$ for \emph{many/all} $(\Sbb,\Tbb)$ pairs, \ie
\beq\label{eq:model_to_data_all}
\min_{\thetav} \max_{\phiv} 
\Ebb_{q(\Sbb,\Tbb)}\big[
\Ebb_{q(\xv_{\Sbb\cup\Tbb})} {\log} \sigma[f_{\phiv}(\xv;\Sbb,\Tbb)] + 
\Ebb_{p_{\thetav}(\xv_{\Tbb} | \xv_{\Sbb}) q(\xv_{\Sbb})} {\log} \sigma[-f_{\phiv}(\xv;\Sbb,\Tbb)]
\big],
\eeq
where $q(\Sbb,\Tbb)$ denotes the sampling process of $(\Sbb,\Tbb)$ (see Appendix \ref{appsec:experiment_settings} for an example implementation),
the discriminator $\sigma[f_{\phiv}(\xv;\Sbb,\Tbb)]$ is also constructed as a ViT, and the optimal $f_{\phi^{*}}(\xv;\Sbb,\Tbb) = {\log} \frac{q(\xv_{\Sbb\cup\Tbb})}{p_{\thetav}(\xv_{\Tbb} | \xv_{\Sbb}) q(\xv_{\Sbb})} = {\log} \frac{q(\xv_{\Tbb} | \xv_{\Sbb})}{p_{\thetav}(\xv_{\Tbb} | \xv_{\Sbb})}$.

In addition to \eqref{eq:model_to_data_all}, we further introduce training tasks by considering that any two model distributions $p_{\thetav}(\xv_{\Tbb_1} | \xv_{\Sbb_1}) q(\xv_{\Sbb_1})$ and $p_{\thetav}(\xv_{\Tbb_2} | \xv_{\Sbb_2}) q(\xv_{\Sbb_2})$ with $\Sbb^{1}\cup\Tbb^{1} = \Sbb^{2}\cup\Tbb^{2}$ should be close to each other.
Accordingly, we enable ``communications'' among any two functionalities of the one universal model $p_{\thetav}(\xv_{\Tbb} | \xv_{\Sbb})$ via
\beq\label{eq:model_to_model_all}
	\min_{\thetav} \max_{\phiv} 
	\Ebb_{q(\Sbb^{1},\Tbb^{1})q(\Sbb^{2},\Tbb^{2})}
	\bigg[
	\bali
	& \Ebb_{p_{\thetav}\!(\xv_{\Tbb^{1}} | \xv_{\Sbb^{1}}) q(\xv_{\Sbb^{1}})} {\log} \sigma[f_{\phiv}\!(\xv;\Sbb^{2},\Tbb^{2}) \!-\! f_{\phiv}\!(\xv;\Sbb^{1},\Tbb^{1})] 
	\\
	& \!+\! \Ebb_{p_{\thetav}\!(\xv_{\Tbb^{2}} | \xv_{\Sbb^{2}}) q(\xv_{\Sbb^{2}})} {\log} \sigma[f_{\phiv}\!(\xv;\Sbb^{1},\Tbb^{1}) - f_{\phiv}\!(\xv;\Sbb^{2},\Tbb^{2})]
	\eali \bigg],
\eeq
where the ``communication'' discriminator can be implicitly constructed with the same neural network $f_{\phiv}(\xv;\Sbb,\Tbb)$ from \eqref{eq:model_to_data_all}. Proofs are given in Appendix \ref{appsec:GAN_derivations}.

Combining \eqref{eq:model_to_data_all} and \eqref{eq:model_to_model_all} yields an adversarial-learning implementation of the unsupervised big learning, which is the first principled adversarial pretraining strategy for foundation models, to our knowledge.

\subsubsection{Maximum-Likelihood Implementation}
\label{sec:ML_implementation}

Consider applications with discrete observations $\xv \in \Zbb^{L\times 1}$, where, \eg $\xv$ denotes a sequence of text words or vector-quantified image patches \cite{ramesh2021zero}.
Equation \eqref{eq:uni_model} of Theorem \ref{theom:unsupervised_biglearn} motivate us to model the distribution $p_{\thetav}(\xv_{\Tbb} | \xv_{\Sbb})$ of multiple target words $\xv_{\Tbb}$ conditioned on source words $\xv_{\Sbb}$, which is challenging considering the correlations among $\xv_{\Tbb}$-words.

One solution to that challenge to leverage the causal LM to auto-regressively construct the model $p_{\thetav}(\xv_{\Tbb} | \xv_{\Sbb})$ conditioned on $\xv_{\Sbb}$, considering only the forward prediction ordering within $\xv_{\Tbb}$.
The core idea of unsupervised big learning, \ie to exhaustively exploit data information, motivate us to step further and learn from various prediction orderings within $\xv_{\Tbb}$\footnote{
	The GAN implementation with \eqref{eq:model_to_data_all} and \eqref{eq:model_to_model_all} need not consider the ordering of $\Tbb$ thanks to its (conditionally) joint modeling nature.
}, which is actually quite similar to the permutation LM \cite{yang2019xlnet} (see Table \ref{tab:BigLearn_special_cases}).

With a Transformer-based universal model $p_{\thetav}(\xv_{\bar\Tbb}|\xv_{\bar\Sbb})$ modeling the generative process of a target word $\xv_{\bar\Tbb}$ given source words $\xv_{\bar\Sbb}$ for \emph{any} $(\bar\Sbb,\bar\Tbb)$ pair, the unsupervised/uni-modal big learning yields
\beq\label{eq:ML_implementation}
	\max_{\thetav} \Ebb_{q(\Sbb,\Tbb)} \sum\nolimits_{(\bar\Sbb,\bar\Tbb) \in \Xi_{\Sbb,\Tbb}} \Ebb_{q(\xv_{\bar\Tbb}|\xv_{\bar\Sbb})} \log  p_{\thetav}(\xv_{\bar\Tbb}|\xv_{\bar\Sbb}),
\eeq
where $q(\Sbb,\Tbb)$ denotes the sampling process of $(\Sbb,\Tbb)$ with random permutations, $\Tbb=\{t_1, t_2, \cdots\}$, $\Xi_{\Sbb,\Tbb} = \{(\Sbb,t_1), (\{\Sbb,t_1\},t_2), (\{\Sbb,t_1, t_2\},t_3), \cdots\}$, often $p_{\thetav}(\xv_{\bar\Tbb}|\xv_{\bar\Sbb}) =\text{Categorical}(\xv_{\bar\Tbb}|\pv_{\thetav}(\xv_{\bar\Sbb}))$ is modeled as a categorical distribution with probabilities $\pv_{\thetav}(\xv_{\bar\Sbb})$, and $\xv_{\bar\Tbb}$ always contain one word.

After unsupervised big learning, $p_{\thetav}(\xv_{\bar\Tbb}|\xv_{\bar\Sbb})$ naturally brings versatile generation and data completion capabilities \wrt \emph{any} predicting order.

\subsection{Big Learning in General Settings}
\label{sec:big_learning}


Thanks to the modeling flexibility of the unsupervised/uni-modal big learning with $\xv \in \Rbb^{L\times D}$, to generalize it into the big learning in general settings, where $\Xv = (\yv, \xv)$ contains an additional supervision $\yv \in \Rbb^{L^{\yv}\times D^{\yv}}$, is straight-forward. 
The key idea is to interpret paired multi-modal data as a ``larger'' sample.

\begin{theorem}[Big learning]
	\label{theom:biglearn}
	With the general/multi-modal setup, where a data sample $\Xv = (\yv, \xv)$\footnote{
		We present with two modalities for simplicity; the presented big learning can be readily generalized to situations with multiple paired modalities.
	} contains both feature $\xv \in \Rbb^{L\times D}$ and its paired supervision $\yv \in \Rbb^{L^{\yv}\times D^{\yv}}$ with the $\Xv$-length index set $\Lbb' = [\Lbb^{\yv},\Lbb]$, its any two {non-overlapping} source/target index subsets $\Sbb' = [\Sbb^{\yv}, \Sbb]$ and $\Tbb' = [\Tbb^{\yv}, \Tbb]$ with $\Sbb' \subset \Lbb'$, $\Tbb' \subseteq \Lbb'$, and $\Tbb' \neq \emptyset$, big learning leverages a universal foundation model $p_{\thetav}(\Xv_{\Tbb'}|\Xv_{\Sbb'})$ (see Fig. \ref{fig:arch_big_learning}) to model many/all joint/conditional/marginal $\Xv$-data distributions simultaneously, \ie 
	\beq\bali\label{eq:multi_model}
		& p_{\thetav}(\Xv_{\Tbb'}|\Xv_{\Sbb'}) \longrightarrow q(\Xv_{\Tbb'}|\Xv_{\Sbb'}), \forall ({\Sbb'},{\Tbb'}) \in {\Omegamat'}
	\eali\eeq
	where ${\Omegamat'}$ can be flexibly specified as the set that contains all $({\Sbb'},{\Tbb'})$ pairs or a portion of them.
	$q(\Xv) \triangleq q(\yv,\xv)$ is the underlying complete data distribution.
	For any $({\Sbb'},{\Tbb'})$, $q(\Xv_{\Tbb'}|\Xv_{\Sbb'})$ is the corresponding joint/conditional/marginal $\Xv$-data distribution, whose samples are readily selected from the whole available data.
	Similar to Theorem \ref{theom:unsupervised_biglearn}, the objective that measures the distance/divergence between both sides of \eqref{eq:multi_model} is application-dependent.
\end{theorem}

\begin{remark}
	For situations where $\Xv = (\yv, \xv)$ has the same data type (\eg both $\yv$ and $\xv$ denote a sequence of \emph{continuous} image-patches), big learning works basically the same as its unsupervised/uni-modal simplification in Theorem \ref{theom:unsupervised_biglearn}.
	However, for challenging situations where each modality has a different data type, \eg where $\yv$ denotes \emph{discrete} text words but $\xv$ are \emph{continuous} image-patches \cite{gupta2021towards,li2021towards,ramesh2021zero,ramesh2022hierarchical,baevski2022data2vec}, one may resort to the following two techniques to enjoy the big learning.
	\begin{enumerate}[leftmargin=5mm]
		\item \textbf{To transform one data type into the other type for alignment}, \eg one can vector-quantize the \emph{continuous} $\xv$ into a sequence of \emph{discrete} tokens \cite{ramesh2021zero}.
		
		\item \textbf{To recursively reuse $p_{\thetav}(\Xv_{\Tbb'}|\Xv_{\Sbb'})$ to isolate each type}, \ie one can unfold the learning via 
		\beq\bali\label{eq:biglearn_doubleFP}
			p_{\thetav}(\Xv_{\Tbb'}|\Xv_{\Sbb'}) 
			= p_{\thetav}(\yv_{\Tbb^{\yv}} |\xv_{\Tbb},\Xv_{\Sbb'})
			p_{\thetav}(\xv_{\Tbb} |\Xv_{\Sbb'})
			= p_{\thetav}(\Xv_{\Tbb^{\yv}} |\Xv_{\Tbb\cup\Sbb'})
			p_{\thetav}(\Xv_{\Tbb} |\Xv_{\Sbb'}),
		\eali\eeq
		where $\Xv_{\Tbb^{\yv}} / \Xv_{\Tbb}$ has one unique data type after unfolding. One can then resort to big learning both $p_{\thetav}(\Xv_{\Tbb^{\yv}} |\Xv_{\Tbb\cup\Sbb'}) \longrightarrow q(\Xv_{\Tbb^{\yv}} |\Xv_{\Tbb\cup\Sbb'})$ and $p_{\thetav}(\Xv_{\Tbb} |\Xv_{\Sbb'}) \longrightarrow q(\Xv_{\Tbb} |\Xv_{\Sbb'})$ for training.		
	\end{enumerate}

\end{remark}

\begin{remark}
	\vspace{1mm}
	From the learning perspective, big learning unifies (and contains as special cases) many machine learning paradigms within one framework, as detailed in Table \ref{tab:BigLearn_special_cases}.
	That generalizability of big learning, combined with its data/task flexibilities, may enable flexible combinations and communications among different learning paradigms.
	
\end{remark}

\subsection{Discussions on Big Learning}
\label{sec:remarks_biglearn}

Without loss of generality, we focus on the simplified notations in the unsupervised settings for presentation and only employ the complicated ones in Theorem \ref{theom:biglearn} if necessary.

\textbf{Can we share one universal foundation model $p_{\thetav}(\xv_{\Tbb}|\xv_{\Sbb})$ among all $(\Sbb,\Tbb)$ pairs? Yes, and it's what we should do.}
All conditional/marginal distributions $q(\xv_{\Tbb}|\xv_{\Sbb})$ can be analytically derived from the joint one $q(\xv)$, meaning that they all share the same set of underlying ``parameters''; 
accordingly, the corresponding joint/conditional/marginal modelings are also expected to share parameters.
Besides, sharing parameters also enables self-regularization among joint/conditional/marginal modelings, which likely encourages model parameters to approach that underlying ``parameters.''


\textbf{On big-learned model parameters and latent features.} 
As aforementioned, most exiting foundation models that exhibit extraordinary robustness, adaptability, and generalization are special cases of the big learning. 
Accordingly, we try to explain from the big learning perspective why they have such amazing characteristics. 
\begin{itemize}[leftmargin=5mm]
	
	\item Firstly, by referring to \eqref{eq:uni_model} and \eqref{eq:multi_model}, both the model parameters and its latent features are shared among many/all joint/conditional/marginal data modeling tasks, which have the same consistent goal of modeling the intrinsic data information (\ie the aforementioned underlying ``parameters'' of $q(\xv_{\Tbb}|\xv_{\Sbb})$) from diverse perspectives.
	Therefore, it's expected that big learning would encourage summarizing intrinsic compositional data knowledge \cite{wu2021lime,lu2021pretrained} in the model parameters (and its latent features), manifested as those amazing characteristics.
	
	\item Secondly, the extraordinary data and task flexibilities of big learning enable large-scale training with massive complete/incomplete data and diverse tasks (across potentially many domains).
	The significantly expanded training experiences (associated with both data and tasks) are expected to effectively reduce the training-test (or pretraining-finetuning) gap and therefore improve the robustness/generalization of big-learned foundation models.
	
\end{itemize}

\textbf{Big learning versus self-supervised contrastive learning.} 
Contrastive learning focuses on exploiting domain prior knowledge to learn generally applicable data representations for downstream tasks \cite{he2020momentum,chen2020simple,grill2020bootstrap,chen2021exploring}. 
From the perspective of prior exploitation, contrastive learning is orthogonal to the big learning that is mostly data-driven. 
One can of course consider leveraging the flexibility of big learning to combine it with contrastive learning to incorporate trustworthy domain priors.

\vspace{-2mm}
\section{Experiments}
\label{sec:Exp}
\vspace{-2mm}

\begin{figure}[tb]
	\vspace{-2mm}
	\centering
	\includegraphics[width=\columnwidth]{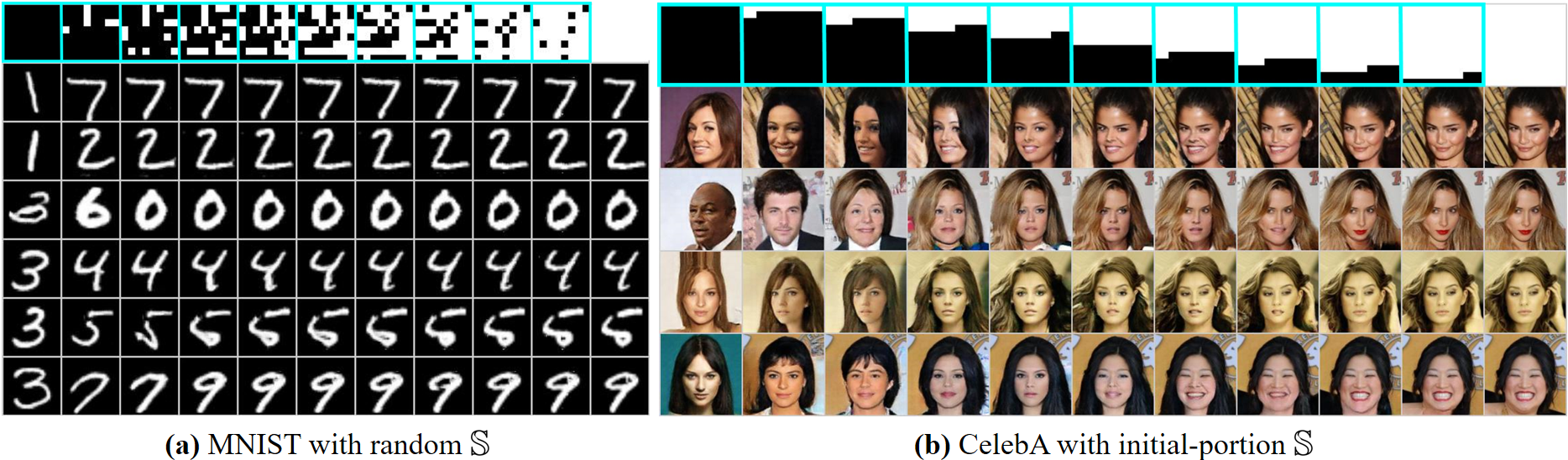}
	\caption{Versatile data generation/completion capabilities from big learning. The first row with light-blue boxes shows different $\Sbb$s, with an increasing $\Sbb$-ratio from left to right. The rightmost column gives the real image.
	}
	\label{fig:increase_Sratio}
		\vspace{-3mm}
\end{figure}

The data/task flexibilities of the big learning significantly expand its scope of application, which, however, also brings tremendous challenges to comprehensive evaluations of its full potential.

Due to our limited computation budget, we concentrate on demonstrating the characteristics of the big learning from diverse perspectives. 
Specifically, we first reveal that unsupervised big learning is indeed capable of delivering \emph{all} joint/conditional/marginal data capabilities, leveraging the adversarial-learning implementation with \eqref{eq:model_to_data_all} and \eqref{eq:model_to_model_all} on the MNIST/CelebA datasets (see Appendix \ref{appsec:experiment_settings} for details).
We then demonstrate the generalization capability of that big-learned foundation model with diverse abused out-of-domain challenges.
Next, based on the maximum-likelihood implementation in \eqref{eq:ML_implementation}, we show that big learning can naturally handle multi-modal data and its joint/conditional/marginal data capabilities directly manifest as versatile functions of great practical value, \eg classification and generation.
Finally, considering the quantitative evaluations of the big learning, we conduct experiments on the GLUE benchmark to reveal that big learning can serve as a superior fine-tuning strategy than the naive one.

It's worth noting that current big/foundation models---the majority of which are the special cases of the big learning---have offered copious evidence for the effectiveness of the big learning, from a variety of special-case perspectives.
Moreover, the co-occurrent work \cite{bao2023one} has provided concrete empirical proofs that fitting all joint/conditional/marginal distributions with one (diffusion) model (\ie performing the big learning) could deliver SOTA performance in representative tasks like text-to-image generation.

\subsection{Versatile Data Completion Capabilities With Adaptive Generation Diversity}
\label{sec:unsupervised_exp_Completion}

We first test the big-learned data generation/completion capabilities with different ratios $r_{\Sbb}$ of $\Sbb$ in $\Lbb$. For a specific $r_{\Sbb}$, we either randomly sample $r_{\Sbb} L$ image patches or choose the first $r_{\Sbb}$-portion to form the source $\xv_{\Sbb}$, which is then input to the model $p_{\thetav}(\xv_{\Tbb}|\xv_{\Sbb})$ for image completion. 

Fig. \ref{fig:increase_Sratio} shows the corresponding results.
It's clear that the big-learned model masters many/all joint/conditional/marginal data capabilities simultaneously. 
Besides, big learning also learns from the data an adaptive generation diversity conditioned on $\xv_{\Sbb}$. 
Specifically, with increasing/decreasing $r_{\Sbb}$ (\ie more/less source information), big learning delivers increasingly deterministic/diverse generations controlled by $\xv_{\Sbb}$/random-noise, following our intuition (see Appendix \ref{appsec:add_exp_results} for more results).

\begin{figure}[tb]
	\centering
	\includegraphics[width=0.8\columnwidth]{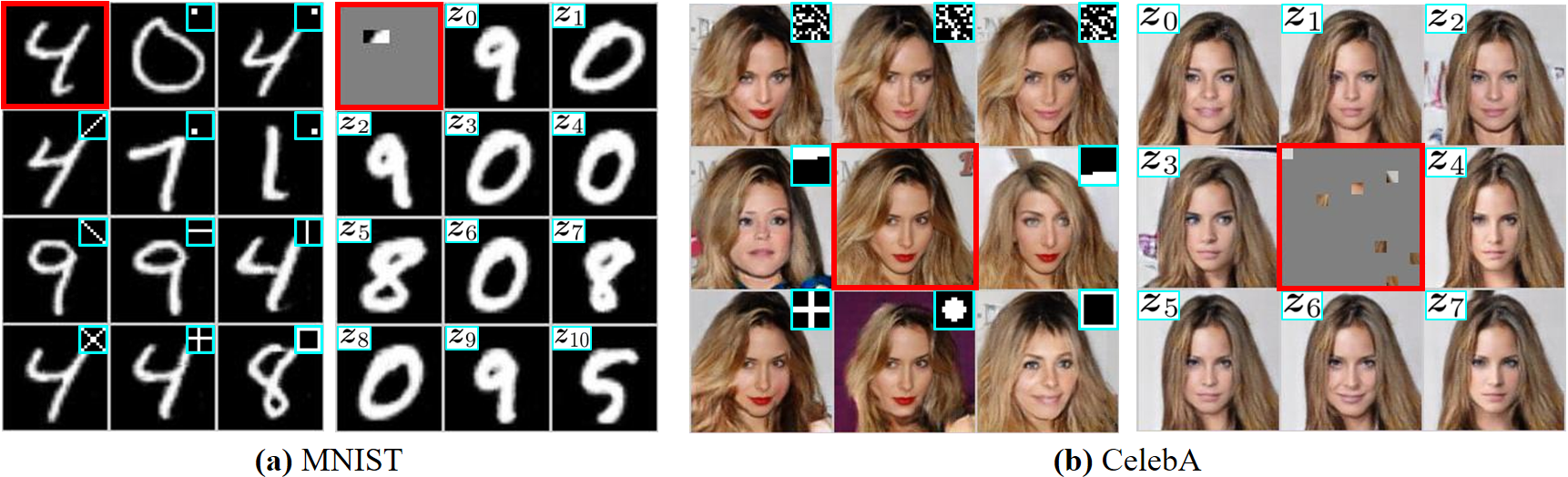}
	\caption{Versatile data completion capabilities from big learning \wrt various $\Sbb$ (left) and noise $\zv$ (right). 
		$\Sbb$s are shown in upper-right light-blue boxes, while the red boxes show $\xv$ (left) and $\xv_{\Sbb}$ (right), respectively. 
	}
	\label{fig:diff_S_noise}
\end{figure}

We then test the big-learned capabilities with respect to various $\Sbb$ and noise settings, with the results summarized in Fig. \ref{fig:diff_S_noise}.
On the one hand, given an image $\xv$ and a random noise $\zv$, big learning clearly delivers for various $\Sbb$s diverse realistic generations on both MNIST (see the variations in class/stroke-thickness/shape/angle) and CelebA (see the varying identity/hair-style/make-up/expression).
On the other hand, given a specific $\xv_{\Sbb}$ with limited information, the big-learned model, when input different noises $\zv_i$, also generates realistic images with diversity.

The experimental results in Figs. \ref{fig:increase_Sratio} and \ref{fig:diff_S_noise} demonstrate that, by comprehensively exploiting the available information inherent in large-scale complete/incomplete data, big learning is capable of delivering versatile data generation/completion capabilities with learned adaptive generation diversity.

\subsection{Generalization on Abused Anomalous Out-Of-Domain Completion}
\label{sec:unsupervised_exp_abuse}

\begin{figure}[tb]
	\centering
	\includegraphics[width=0.95\columnwidth]{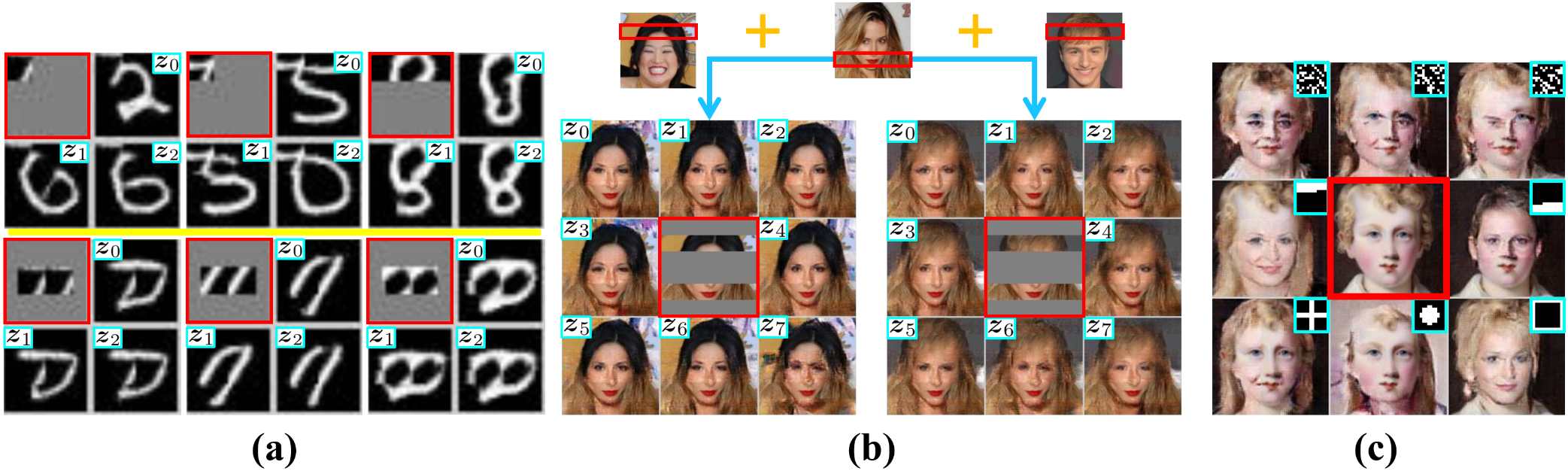}
	\caption{Abused anomalous completion for demonstrating the generalization of big learning.
		(a) $\xv_{\Sbb}$ constructed with random center patches replaced in the upper-left corner (top) and duplicated and replaced in the center (bottom).
		A model big-learned on CelebA is used in (b)-(c).
		(b) $\xv_{\Sbb}$ combining patches from different images.
		(c) Out-of-domain $\xv_{\Sbb}$ from MetFaces \cite{karras2020training}.
	}
	\label{fig:abused_testing}
\end{figure}

We design abused completion tasks to test the generalization of the big learning. 
Specifically, we intentionally design $\xv_{\Sbb}$ with 
($i$) abused interventions to source patches (\eg random relocation and duplication, as shown in Fig. \ref{fig:abused_testing}(a)); 
($ii$) mixed-up patches from different data samples (see Fig. \ref{fig:abused_testing}(b));
and ($iii$) unseen out-of-domain image patches, as shown in Fig. \ref{fig:abused_testing}(c).

It's clear that big learning manages to handle these abused $\xv_{\Sbb}$ with reasonable image completion; \eg see the realistic characters with overall consistent style and smooth strokes in Fig. \ref{fig:abused_testing}(a), the harmoniously completed faces even with mismatched face frame and hair color in Fig. \ref{fig:abused_testing}(b), and the relatively smooth out-of-domain completion in Fig. \ref{fig:abused_testing}(c).
These surprising results from abused anomalous out-of-domain completions (along with the great successes of existing foundation models) validate the generalization capability of the presented big learning.  


\subsection{Leveraging Big Learning to Unify Classification and Generation}

\begin{figure}[tb]
	\centering
	\subfloat[Joint Generation]{
		\includegraphics[height=0.23\columnwidth]{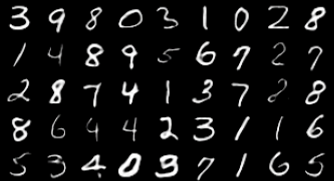}
		\label{fig:}}
	\qquad
	\subfloat[Label-Conditioned Generation ]{
		\includegraphics[height=0.23\columnwidth]{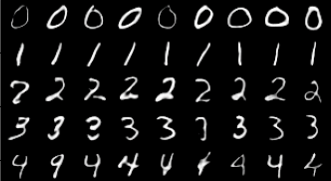}
		\label{fig:}}
	\qquad
	\caption{
		Demonstration of versatile data capabilities of big learning, retrieved from $p_{\thetav}(\Xv_{\Tbb'}|\Xv_{\Sbb'})$ with specified $(\Sbb', \Tbb')$.
	}
	\label{fig:biglearn_genclass}
		\vspace{-2mm}
\end{figure}

We test the big learning in the general settings, where $\Xv = (y, \xv)$ contains both image tokens $\xv \in \Zbb^{L\times 1}$ and a paired label $y \in \{1, \cdots, C\}^{1\times 1}$.
We conduct the experiment on MNIST and follow \cite{bao2021beit,ramesh2021zero} to first vector-quantize an image for its tokens $\xv$, followed by big learning based on \eqref{eq:ML_implementation}.
Details are given in Appendix \ref{appsec:biglearn_genclass}.

Given the big-learned universal model $p_{\thetav}(\Xv_{\Tbb'}|\Xv_{\Sbb'})$, one can retrieve from it versatile data capabilities by specifying the corresponding $(\Sbb', \Tbb')$, such as joint generation (\ie $p_{\thetav}(\xv)$; see Fig. \ref{fig:biglearn_genclass}(a) for the results), label-conditioned generation (\ie $p_{\thetav}(\xv|y)$; see Fig. \ref{fig:biglearn_genclass}(b)), classification (\ie $p_{\thetav}(y|\xv)$), random completion (\ie $p_{\thetav}(\xv_{\Tbb}|\xv_{\Sbb})$), label-conditioned completion (\ie $p_{\thetav}(\xv_{\Tbb}|\xv_{\Sbb}, y)$), \etc
These simultaneously-delivered capabilities are likely valuable for counterfactual analysis and reasoning.

\vspace{-2mm}
\subsection{Quantitative Evaluations on the GLUE Benchmark}

Because of our limited computation budget, we cannot afford to make systematic quantitative comparisons between the big learning and existing methods on pretraining a foundation model with large-scale data. 
Alternatively, we empirically reveal that the big learning is a superior fine-tuning strategy than the naive one.

Specifically, we initialize with the pretrained \texttt{xlnet-base-cased} model from the Hugging Face transformers library \cite{wolf-etal-2020-transformers} and then test fine-tuning it on downstream RTE/MRPC/SST-2 tasks (from the GLUE Benchmark \cite{wang2018glue}) with ($i$) the naive fine-tuning strategy (termed FT) and ($ii$) the big learning (termed big-learn), respectively.
Table \ref{apptab:glue_ACC} summarizes the quantitative evaluation results, where it's clear that big-learn consistently outperforms FT, even without careful tuning. See Appendix \ref{appsec:empirical_evaluation} for details.

\begin{table}[tb]
	\centering
	\caption{Big learning serves as a superior fine-tuning strategy.
		The best/median metrics are calculated among the combinations of the tested hyperparameters of Table \ref{apptab:glue_hyperpara}.}
	\vspace{2mm}
		\resizebox{0.6\columnwidth}{!}{
		\begin{tabular}{l | c c | c c}
			\hline \hline
			\multirow{2}{*}{Task} & \multicolumn{2}{c|}{Best Accuracy / F1} & \multicolumn{2}{c}{Median Accuracy / IQR}
			\\ 
			& FT & big-learn & FT & big-learn
			\\ \hline 
			RTE & 71.84  & $\textbf{75.09}$ & 66.06/2.34  & $\textbf{70.75/1.44}$ 
			\\
			MRPC & 88.97/92.09 & $\textbf{90.20/93.03}$ & 87.00/2.45 & $\textbf{87.74/1.10}$  
			\\
			SST-2 & 94.15  & $\textbf{95.18}$ & 93.75/0.45 & $\textbf{94.66/0.28}$  
			\\ \hline \hline
		\end{tabular}
		}
	\label{apptab:glue_ACC}
	\vspace{-2mm}
\end{table}

\vspace{-2mm}
\section{Conclusions}

We propose the big learning that exhaustively exploits the available data information and potentially delivers all joint/conditional/marginal data capabilities.
We reveal that big learning ($i$) comes with extraordinary training flexibilities for complete/incomplete data and for customizing training tasks, ($ii$) contains most existing foundation models as special cases, and ($iii$) unifies conventional machine learning paradigms and enables their flexible cooperations. 
Though inspiring, big learning also shares the same constraints of foundation models \cite{bommasani2021opportunities,yuan2022roadmap}; \eg either to comprehensively reveal its potential or to verify its effectiveness (for general downstream tasks) is extremely challenging and time-consuming.
Therefore, we believe that big learning needs our community and vice versa.


%
%

%

%

{\small
		\bibliography{ReferencesCong}
	\bibliographystyle{plain}
}
\newpage
\appendix
\onecolumn

\begin{center}
    {\large
    \textbf{
    Appendix of Big Learning
    }}
    \vspace{3mm}
    

	
\end{center}

\vskip 0.3in

\section{On naive modeling of all joint/conditional/marginal data distributions}
\label{appsec:naive_model_big_learning}

We present with the unsupervised settings, where $\xv\in \Rbb^{L\times D}$ with length $L$ and dimension $D$ (like $L$ flattened patches of an image or $L$ words with $D=1$). It's straightforward to generalize the following analyses to the general settings with a data sample $\Xv = (\yv, \xv)$ contains an additional supervision $\yv \in \Rbb^{L^{\yv}\times D^{\yv}}$.
Considering $D>1$ and $D=1$ for image patches and text words, respectively, we concentrate on analyzing the modeling of all joint/conditional/marginal data distributions \wrt the length $L$ below.

As mentioned in the main manuscript, one need construct $N_{\text{all}}=\sum_{i=0}^{L-1} C_L^{i} (\sum_{k=1}^{L-i} C_{L-i}^k)$ models to naively model all joint/conditional/marginal data distributions, to collect all joint/conditional/marginal data capabilities.
$C_L^i$ denotes the number of $i$-combinations from a set with $L$ elements.

To elaborate on that, consider a simple $3$-length $1$-dimensional problem with $\xv=[x_1,x_2,x_3]^T$, where $L=3$, $D=1$, $x_i\in \Rbb$, and the length index set $\Lbb=\{1,2,3\}$.
\begin{itemize}
	
	\item The goal of the joint modeling is to deliver $p_{\thetav}(\xv) \longrightarrow q(\xv)$ with one model $p_{\thetav}(\xv)$.
	
	\item By contrast, to naively model all joint/conditional/marginal data distributions, one need construct $19$ models for such a simple $3$-length problem, \ie
	\beq\bali \label{eq:selfcondition_3D}
	& p_{\thetav^{1}}(x_1),  \,  p_{\thetav^{2}}(x_2),   \,  p_{\thetav^{3}}(x_3),   \,
	p_{\thetav^{4}}(x_1, x_2),   \, p_{\thetav^{5}}(x_2,x_3), \, p_{\thetav^{6}}(x_1,x_3),  \,   p_{\thetav^{7}}(x_1, x_2, x_3),
	\\
	& p_{\thetav^{8}}(x_2|x_1),   \,  p_{\thetav^{9}}(x_3|x_1),   \, p_{\thetav^{10}}(x_2,x_3|x_1), 
	\\
	& p_{\thetav^{11}}(x_1|x_2),  \,   p_{\thetav^{12}}(x_3|x_2),  \,  p_{\thetav^{13}}(x_1,x_3|x_2), 
	\\
	& p_{\thetav^{14}}(x_1|x_3),  \, p_{\thetav^{15}}(x_2|x_3),  \, p_{\thetav^{16}}(x_1, x_2|x_3), 
	\\
	& p_{\thetav^{17}}(x_1|x_2,x_3), p_{\thetav^{18}}(x_2|x_1,x_3), p_{\thetav^{19}}(x_3|x_1,x_2).
	\eali\eeq
\end{itemize}

Based on the above $3$-length problem, one can readily summarize the following two steps in calculating the number of models in naively modeling all joint/conditional/marginal data distributions, \ie $q(\xv_{\Tbb}|\xv_{\Sbb}), \forall \Sbb\subset\Lbb, \Tbb\subseteq\Lbb,\Tbb\neq\emptyset$.
\begin{enumerate}
	\item \textbf{Sample $\Sbb$.} The source index set $\Sbb$ may contain $\{0, \cdots, L-1\}$ indexes/locations, where $\Sbb$ containing $0$ index corresponds to joint/marginal generations and $\Sbb$ containing $\ge 1$ indexes corresponds to conditional generations/completions.
	For a special case with $i$ indexes in $\Sbb$ with $i\in[0, L-1]$, one has $C_L^{i}$ ways to specify that source index set $\Sbb$.
	
	\item \textbf{Sample $\Tbb$ conditioned on $\Sbb$.} Given a $\Sbb$ consisting of $i$ indexes, the target index set $\Tbb$ could contain $\{1, \cdots, L-i\}$ indexes/locations outside $\Sbb$. For a special case of $\Tbb$ containing $k$ indexes where $k\in[1, L-i]$, one has $C_{L-i}^{k}$ ways to specify the target $\Tbb$.
	
\end{enumerate}

Therefore, to naively model all joint/conditional/marginal data distributions, one need construct $N_{\text{all}}=\sum_{i=0}^{L-1} C_L^{i} (\sum_{k=1}^{L-i} C_{L-i}^k)$ models, which, however, is prohibitive in practice.

Note with ideal modeling of $q(\xv_{\Tbb}|\xv_{\Sbb})$, the orders in $\Sbb/\Tbb$ should not matter.
However, that may not hold true considering practical constraints, \eg where existing joint modeling techniques fail to model the multi-mode characteristics of $\xv_{\Tbb}$.
Besides, in the NLP application of language modeling, one may be interested in versatile (conditional) generation ordering (as defined in $\Tbb$), mimicking the permutation language modeling \cite{yang2019xlnet}. 
In that case, to naively modeling all joint/conditional/marginal data distributions, one need construct $N_{\text{all}}'=\sum_{i=0}^{L-1} C_L^{i} (\sum_{k=1}^{L-i} A_{L-i}^k)$ models to take into consideration the order of $\Tbb$, where the order of $\Sbb$ is ignored and $A_{L-i}^k$ denotes the number of the ordered arrangements of $k$ elements from a set with $L-1$ elements.
Similarly, one need construct $N_{\text{all}}''=\sum_{i=0}^{L-1} A_L^{i} (\sum_{k=1}^{L-i} A_{L-i}^k)$ models to model the orders in both $\Sbb$ and $\Tbb$.


\section{Derivations of the GAN example associated with Eqs. \eqref{eq:model_to_data_all} and \eqref{eq:model_to_model_all}}
\label{appsec:GAN_derivations}

Here we present the detailed derivations/proofs for the GAN example associated with Eqs. \eqref{eq:model_to_data_all} and \eqref{eq:model_to_model_all} of the main manuscript. 
For better understanding, we begin with a simplified case where $\Tbb = \Lbb \backslash \Sbb$, followed by generalizing the results to the general situations with $\Tbb \subseteq \Lbb \backslash \Sbb$.

\subsection{$\bb \Tbb = \Lbb\backslash\Sbb$}
\label{sec:GAN_joint}

To leverage the GAN training framework \cite{goodfellow2014generative}, one needs the sampling capabilities from the distributions of interest.
With $\Tbb = \Lbb\backslash\Sbb$, here we are interested in the joint distributions with accessible sampling capabilities, including
\beq\bali\label{appeq:samplablePDFs_Joint}
q(\xv) &
\\
p_{\thetav}(\xv;\Sbb) & = p_{\thetav}(\xv_{\Lbb\backslash\Sbb} | \xv_{\Sbb}) q(\xv_{\Sbb}) \quad \forall \Sbb.
\eali\eeq
Note one can of course exploit the flexibility of big learning to define other joint distributions with sampling capabilities, such as an recursively defined distribution
\beq
p_{\thetav}(\xv;\Sbb^1,\Sbb^2) = p_{\thetav}(\xv_{\Lbb\backslash\Sbb^2} | \xv_{\Sbb^2}) p_{\thetav}(\xv_{\Sbb^2}), 
\eeq
where $p_{\thetav}(\xv_{\Sbb^2}) = \int p_{\thetav}(\xv_{\Lbb\backslash\Sbb^1} | \xv_{\Sbb^1}) q(\xv_{\Sbb^1}) d \xv_{\Lbb\backslash\Sbb^2}$. 
For simplicity, we focus on the simplified settings in Eq. \eqref{appeq:samplablePDFs_Joint} and leave the interesting but complicated recursive case for future research.

\begin{figure}
	\centering
	\subfloat[Case 1]{
		\includegraphics[height=0.25\columnwidth]{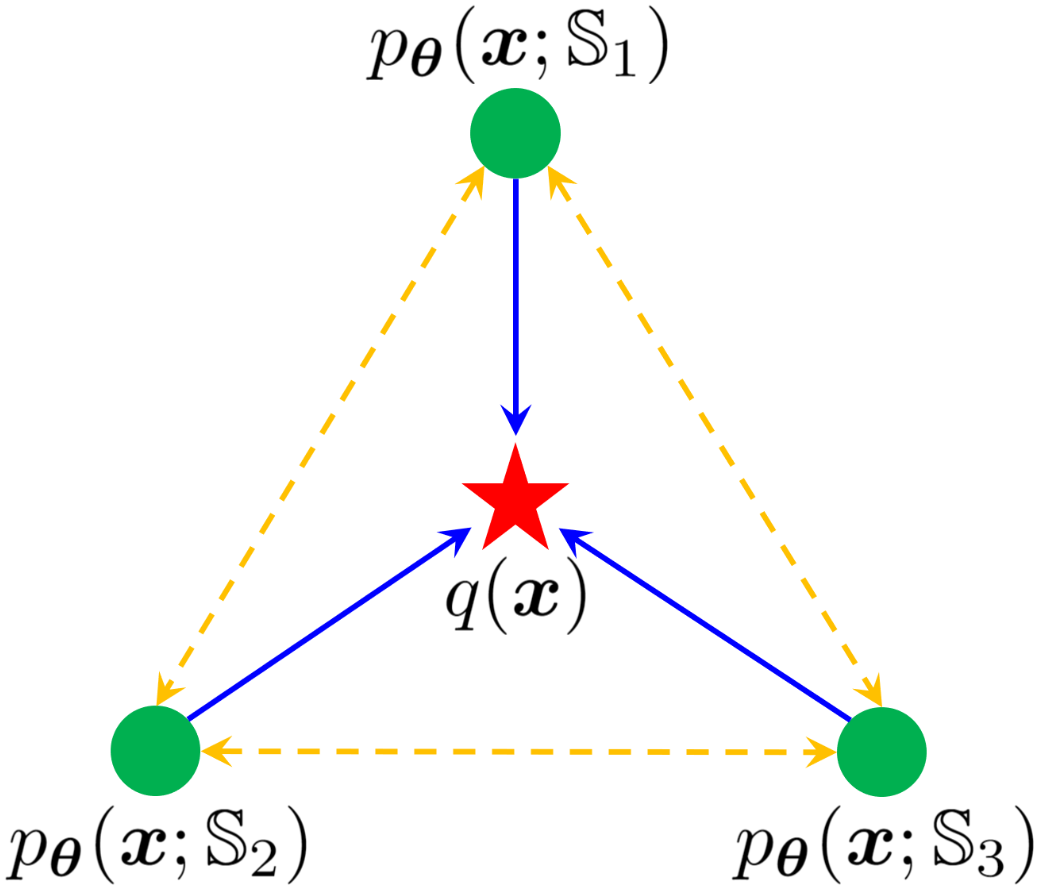}
		\label{fig:}}
	\qquad
	\subfloat[Case 2]{
		\includegraphics[height=0.25\columnwidth]{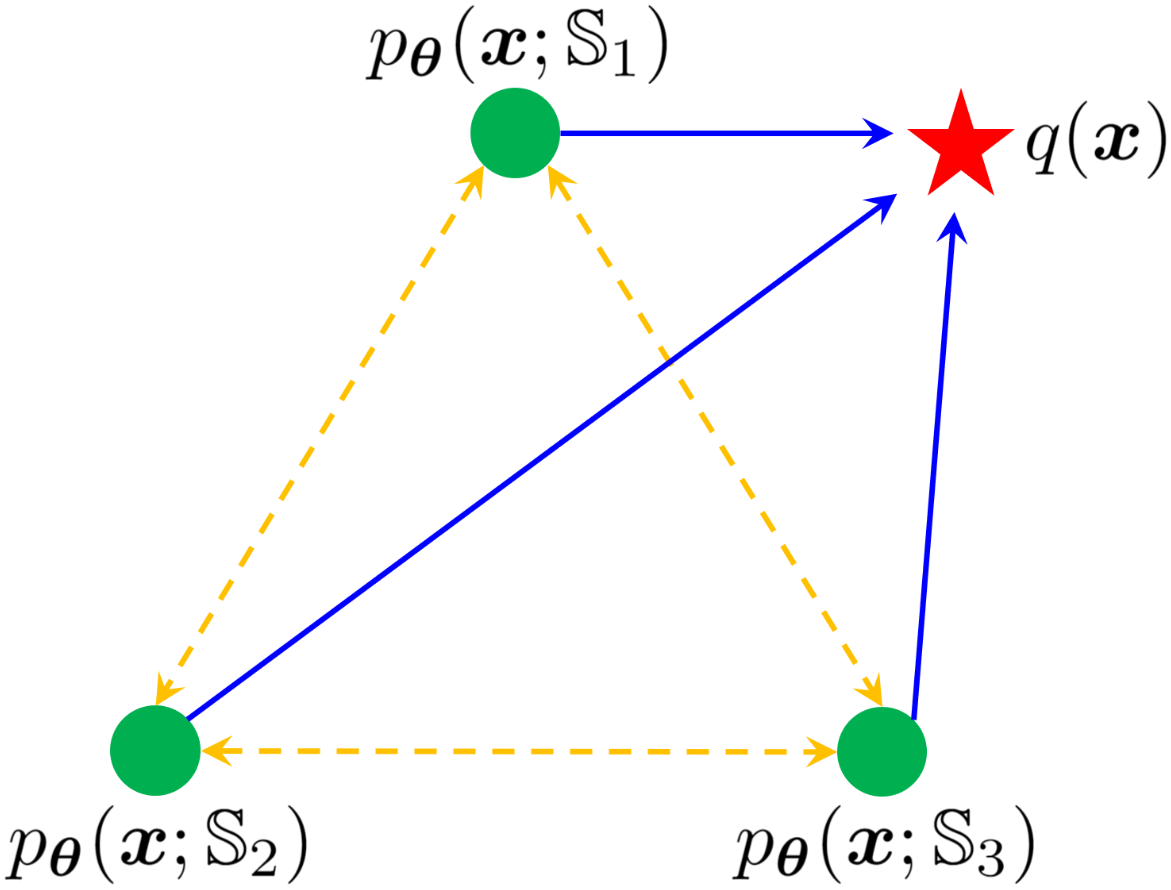}
		\label{fig:}}
	\caption{Demonstration of unsupervised big learning based on GANs.}
	\label{fig:demo_unsupervised_biglearn_GAN}
\end{figure}

Given the underlying data distribution $q(\xv)$ and ``model'' distributions $p_{\thetav}(\xv;\Sbb)$ in Eq. \eqref{appeq:samplablePDFs_Joint}, 
\begin{enumerate}
    \item one can match any $p_{\thetav}(\xv;\Sbb)$ to $q(\xv)$ adversarially with a GAN. Take the standard GAN \cite{goodfellow2014generative} for an example, the objective is
    \beq\bali\label{appeq:model_to_data_joint}
    \min_{\thetav} \max_{\phiv} \Ebb_{q(\xv)} \log \sigma(f_{\phiv}(\xv; \Sbb)) + \Ebb_{p_{\thetav}(\xv_{\Lbb\backslash\Sbb} | \xv_{\Sbb}) q(\xv_{\Sbb})} \log (1-\sigma(f_{\phiv}(\xv; \Sbb))),
    \eali\eeq
    where the optimal $f_{\phi^{*}}(\xv; \Sbb) = \log \frac{q(\xv)}{p_{\thetav}(\xv_{\Lbb\backslash\Sbb} | \xv_{\Sbb}) q(\xv_{\Sbb})} = \log \frac{q(\xv_{\Lbb\backslash\Sbb} | \xv_{\Sbb})}{p_{\thetav}(\xv_{\Lbb\backslash\Sbb} | \xv_{\Sbb})}$. 
    Ideally, optimizing the above objective is identical to minimizing the Jensen-Shannon divergence $\JS[q(\xv)||p_{\thetav}(\xv;\Sbb)]$, as illustrated with the blue solid arrows in Fig. \ref{fig:demo_unsupervised_biglearn_GAN}.
    
    \item one can also conduct matching among any two model distributions (\eg $p_{\thetav}(\xv;\Sbb^1)=p_{\thetav}(\xv_{\Lbb\backslash\Sbb^1} | \xv_{\Sbb^1}) q(\xv_{\Sbb^1})$ and $p_{\thetav}(\xv;\Sbb^2)=p_{\thetav}(\xv_{\Lbb\backslash\Sbb^2} | \xv_{\Sbb^2}) q(\xv_{\Sbb^2})$) to enable communications/cooperations among them, via optimizing
    \beq\label{appeq:model_to_model_joint}
    \min_{\thetav} \max_{\phiv} 
    \left\{\bali
    & \Ebb_{p_{\thetav}(\xv_{\Lbb\backslash\Sbb^1} | \xv_{\Sbb^1}) q(\xv_{\Sbb^1})} {\log} \sigma(f^{'}_{\phiv}(\xv;\Sbb^{1},\Sbb^{2})) 
    \\
    & + \Ebb_{p_{\thetav}(\xv_{\Lbb\backslash\Sbb^2} | \xv_{\Sbb^2}) q(\xv_{\Sbb^2})} {\log} (1-\sigma(f^{'}_{\phiv}(\xv;\Sbb^{1},\Sbb^{2})))
    \eali
    \right.
    \eeq
    where the optimal $f'_{\phi^{*}}(\xv;\Sbb^{1},\Sbb^{2}) = \log \frac{p_{\thetav}(\xv_{\Lbb\backslash\Sbb^1} | \xv_{\Sbb^1}) q(\xv_{\Sbb^1})}{p_{\thetav}(\xv_{\Lbb\backslash\Sbb^2} | \xv_{\Sbb^2}) q(\xv_{\Sbb^2})}$.
    The orange dotted arrows in Fig. \ref{fig:demo_unsupervised_biglearn_GAN} demonstrate such idea.
        
\end{enumerate}

At first sight of Eqs. \eqref{appeq:model_to_data_joint} and \eqref{appeq:model_to_model_joint}, it seems one should at least construct two discriminators, with $f_{\phi}(\xv;\Sbb)$ and $f'_{\phi}(\xv;\Sbb^{1},\Sbb^{2})$ respectively. 
However, we notice that 
$$\bali
f'_{\phi^{*}}(\xv;\Sbb^{1},\Sbb^{2}) 
& = \log \frac{q(\xv)}{p_{\thetav}(\xv_{\Lbb\backslash\Sbb^2} | \xv_{\Sbb^2}) q(\xv_{\Sbb^2})} - \log \frac{q(\xv)}{p_{\thetav}(\xv_{\Lbb\backslash\Sbb^1} | \xv_{\Sbb^1}) q(\xv_{\Sbb^1})}
\\
& = f_{\phi^{*}}(\xv; \Sbb^{2}) - f_{\phi^{*}}(\xv; \Sbb^{1}).
\eali$$
Accordingly, we propose to employ further simplification that builds $f'_{\phi}(\xv;\Sbb^{1},\Sbb^{2})$ on top of $f_{\phi}(\xv;\Sbb)$, \ie we reformulate Eq. \eqref{appeq:model_to_model_joint} as 
\beq\label{appeq:model_to_model_joint_v1}
	\min_{\thetav} \max_{\phiv} 
	\left\{\bali
	& \Ebb_{p_{\thetav}(\xv_{\Lbb\backslash\Sbb^1} | \xv_{\Sbb^1}) q(\xv_{\Sbb^1})} {\log} \sigma[f_{\phi}(\xv; \Sbb^{2}) - f_{\phi}(\xv; \Sbb^{1})] 
	\\
	& + \Ebb_{p_{\thetav}(\xv_{\Lbb\backslash\Sbb^2} | \xv_{\Sbb^2}) q(\xv_{\Sbb^2})} {\log} \sigma[f_{\phi}(\xv; \Sbb^{1}) - f_{\phi}(\xv; \Sbb^{2})].
	\eali\right.
\eeq



Till now, we present the derivations associated with $\Tbb = \Lbb\backslash\Sbb$, \ie matching in the joint space. 
In what follows, we generalize to the settings with $\Tbb \subseteq \Lbb\backslash\Sbb$, to deliver (unsupervised) big learning in all joint/conditional/marginal spaces.

\subsection{$\bb \Tbb \subseteq \Lbb\backslash\Sbb$}
\label{sec:GAN_all}

Similar to the previous section, we also consider simplified situations with no recursiveness, that is, we do not consider a model distribution $p_{\thetav}(\xv_{\Tbb} | \xv_{\Sbb}) p_{\thetav}(\xv_{\Sbb})$, even though such recursive flexibility of big learning is quite interesting.
We leave that as future research.

Accordingly, the considered joint/conditional/marginal distributions with sampling capabilities are
\beq\bali\label{appeq:samplablePDFs_All}
q(\xv_{\Sbb \cup \Tbb})   & 
\\
p_{\thetav}(\xv_{\Sbb \cup \Tbb}) & = p_{\thetav}(\xv_{\Tbb} | \xv_{\Sbb}) q(\xv_{\Sbb}) & \quad \forall  \Sbb, \Tbb
\eali\eeq
where $\Sbb \cup \Tbb$ need not be $\Lbb$. Note $\Sbb \cup \Tbb \subset \Lbb$ means the corresponding $q(\xv_{\Sbb \cup \Tbb})$ is a \emph{marginal} data distribution, whose data samples are readily accessible from those of $q(\xv)$.

Similar to the previous section,
\begin{itemize}
    \item one can match any model distribution $p_{\thetav}(\xv_{\Sbb \cup \Tbb})$ to the underlying joint/marginal data distribution $q(\xv_{\Sbb \cup \Tbb})$, via the standard GAN objective
    \beq\bali\label{appeq:model_to_data}
    \min_{\thetav} \max_{\phiv} \Ebb_{q(\xv_{\Sbb\cup\Tbb})} \log \sigma(f_{\phiv}(\xv;\Sbb,\Tbb)) + \Ebb_{p_{\thetav}(\xv_{\Tbb} | \xv_{\Sbb}) q(\xv_{\Sbb})} \log (1-\sigma(f_{\phiv}(\xv;\Sbb,\Tbb))),
    \eali\eeq
    where $f_{\phi^{*}}(\xv;\Sbb,\Tbb) = \log \frac{q(\xv_{\Sbb\cup\Tbb})}{p_{\thetav}(\xv_{\Tbb} | \xv_{\Sbb}) q(\xv_{\Sbb})} = \log \frac{q(\xv_{\Tbb} | \xv_{\Sbb})}{p_{\thetav}(\xv_{\Tbb} | \xv_{\Sbb})}$.
    
    \item one can also conduct matching among any two model distributions, \eg $p_{\thetav}(\xv_{\Tbb^{1}} | \xv_{\Sbb^{1}}) q(\xv_{\Sbb^{1}})$ and $p_{\thetav}(\xv_{\Tbb^{2}} | \xv_{\Sbb^{2}}) q(\xv_{\Sbb^{2}})$, as long as $\Sbb^{1} \cup \Tbb^{1} = \Sbb^{2} \cup \Tbb^{2}$, with the corresponding objective 
    \beq\label{appeq:model_to_model}
    \min_{\thetav} \max_{\phiv} 
    \left\{\bali 
    & \Ebb_{p_{\thetav}(\xv_{\Tbb^{1}} | \xv_{\Sbb^{1}}) q(\xv_{\Sbb^{1}})} {\log} \sigma(f_{\phiv}(\xv;\Sbb^{1},\Tbb^{1},\Sbb^{2},\Tbb^{2})) 
    \\
    & + \Ebb_{p_{\thetav}(\xv_{\Tbb^{2}} | \xv_{\Sbb^{2}}) q(\xv_{\Sbb^{2}})} {\log} (1-\sigma(f_{\phiv}(\xv;\Sbb^{1},\Tbb^{1},\Sbb^{2},\Tbb^{2}))),
    \eali\right.\eeq
    where $f'_{\phi^{*}}(\xv;\Sbb^{1},\Tbb^{1},\Sbb^{2},\Tbb^{2}) = \log \frac{p_{\thetav}(\xv_{\Tbb^{1}} | \xv_{\Sbb^{1}}) q(\xv_{\Sbb^{1}})}{p_{\thetav}(\xv_{\Tbb^{2}} | \xv_{\Sbb^{2}}) q(\xv_{\Sbb^{2}})}$.
    
    For further simplifications, we again resort to 
    $$\bali
    f'_{\phi^{*}}(\xv;\Sbb^{1},\Tbb^{1},\Sbb^{2},\Tbb^{2})
    & = \log \frac{q(\xv_{\Sbb^{2}\cup\Tbb^{2}})}{p_{\thetav}(\xv_{\Tbb^{2}} | \xv_{\Sbb^{2}}) q(\xv_{\Sbb^{2}})}
    - \log \frac{q(\xv_{\Sbb^{1}\cup\Tbb^{1}})}{p_{\thetav}(\xv_{\Tbb^{1}} | \xv_{\Sbb^{1}}) q(\xv_{\Sbb^{1}})}
    \\
    & = f_{\phi^{*}}(\xv;\Sbb^{2},\Tbb^{2}) - f_{\phi^{*}}(\xv;\Sbb^{1},\Tbb^{1})
    \eali$$
    and build $f'_{\phi}(\xv;\Sbb^{1},\Tbb^{1},\Sbb^{2},\Tbb^{2})$ on top of $f_{\phi}(\xv;\Sbb,\Tbb)$.
    
    Accordingly, Eq. \eqref{appeq:model_to_model} is reformulated as 
    \beq\label{appeq:model_to_model_v1}
    \min_{\thetav} \max_{\phiv} 
    \left\{\bali 
	    & \Ebb_{p_{\thetav}(\xv_{\Tbb^{1}} | \xv_{\Sbb^{1}}) q(\xv_{\Sbb^{1}})} {\log} \sigma[f_{\phiv}(\xv;\Sbb^{2},\Tbb^{2}) - f_{\phiv}(\xv;\Sbb^{1},\Tbb^{1})] 
	    \\
	    & + \Ebb_{p_{\thetav}(\xv_{\Tbb^{2}} | \xv_{\Sbb^{2}}) q(\xv_{\Sbb^{2}})} {\log} \sigma[f_{\phiv}(\xv;\Sbb^{1},\Tbb^{1}) - f_{\phiv}(\xv;\Sbb^{2},\Tbb^{2})].
    \eali\right.\eeq
    
\end{itemize}

Accordingly, we conclude the proofs for the GAN example of the main manuscript.

\begin{figure}[H]
	\centering
	\subfloat[Unsupervised Big Learning]{
		\includegraphics[height=0.38\columnwidth]{Figures/SharedCondPDF.png}
		\label{fig:}}
	\qquad
	\subfloat[MAE \cite{he2021masked} and MaskFeat  \cite{wei2021masked}]{
		\includegraphics[height=0.38\columnwidth]{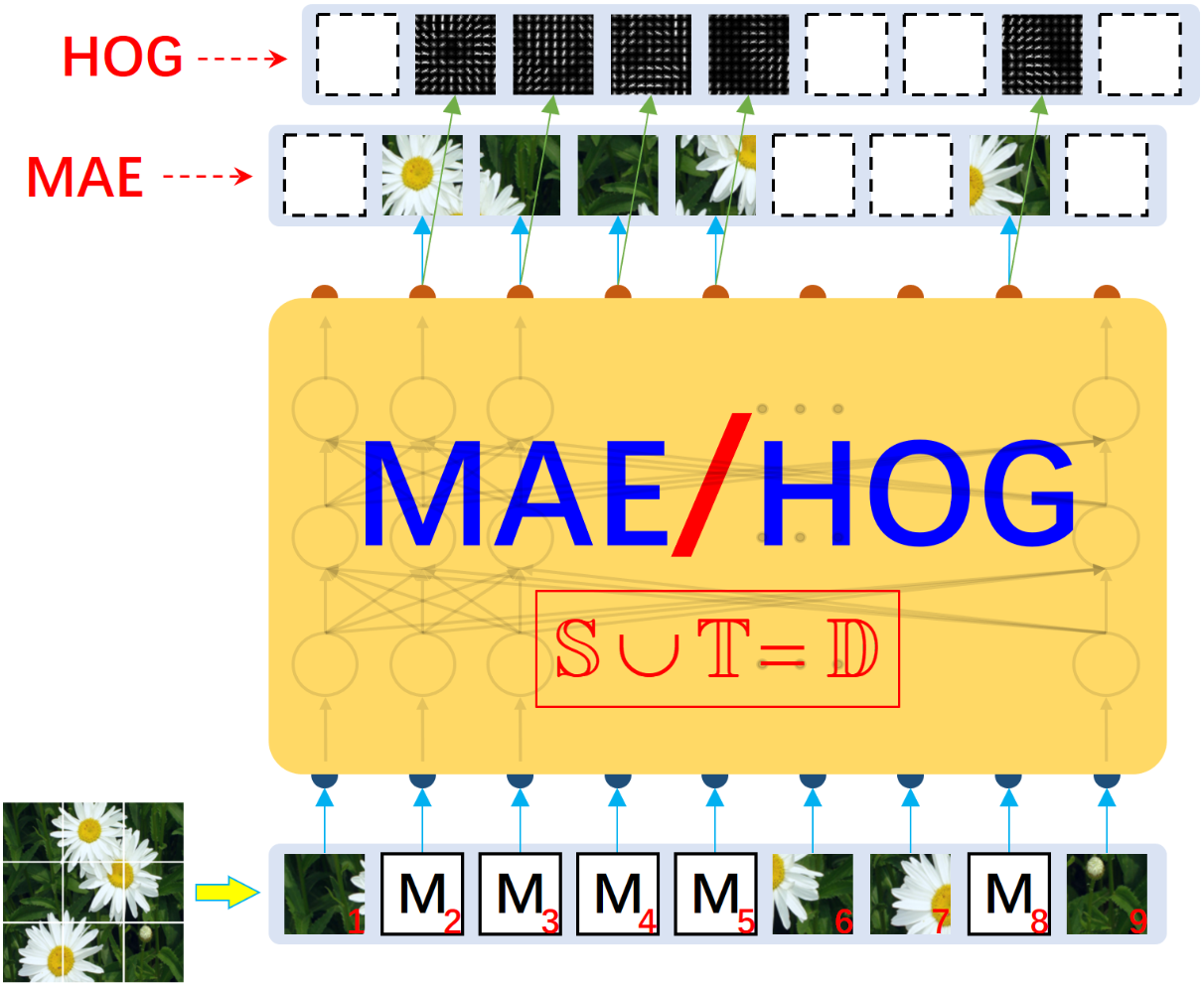}
		\label{fig:arch_MAE_HOG}}
	\caption{Unsupervised big learning (a) and its special cases (b). 
		Often a {mask} token \texttt{[M]} is inserted to the input locations outside $\Sbb$ for forward propagation, while no loss is back-propagated to the output locations outside $\Tbb$. 
		Note inserting the \texttt{[M]} tokens later in a middle layer (but at the same location) often lightens the computation and memory burdens but improves the performance \cite{he2021masked}.
	}
	\label{fig:unsupervised_biglearning}
\end{figure}

\begin{figure}[H]
	\centering
	\subfloat[Big Learning]{
		\includegraphics[height=0.36\columnwidth]{Figures/UnivCondPDF.png}
		\label{fig:}}
	\qquad\quad
	\subfloat[BERT \cite{stickland2019bert}]{
		\includegraphics[height=0.35\columnwidth]{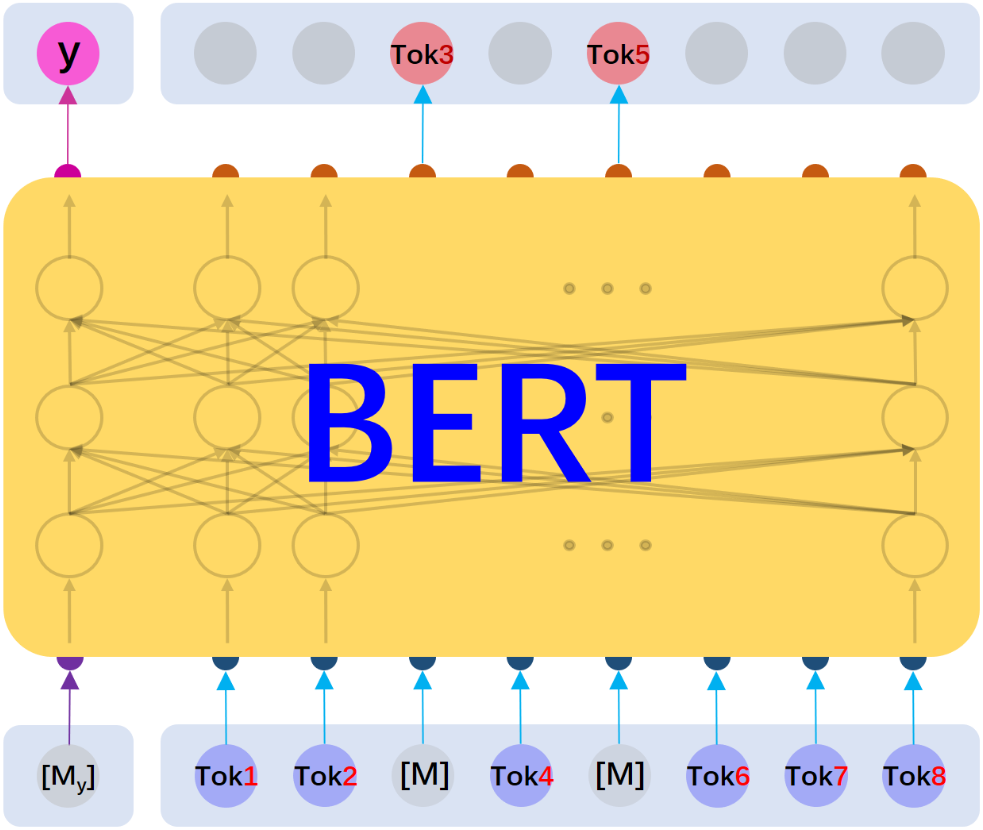}
		\label{fig:special_BERT}}
	\caption{Big learning (a) and its special case of BERT (b). 
		Similar to the {mask} token \texttt{[M]} for $\xv$ (see Fig. \ref{fig:arch_MAE_HOG}), we employ another {mask} token $\texttt{[M}_{\texttt{y}}\texttt{]}$ for $\yv$, which works identically to the classification token \texttt{[CLS]} in BERT settings \cite{stickland2019bert} and the start-of-sentence token in GPT settings \cite{brown2020language}.
		Often inserting \texttt{[M]}/$\texttt{[M}_{\texttt{y}}\texttt{]}$ tokens later in a middle layer improves performance \cite{he2021masked,touvron2021going}.
	}
	\label{fig:big_learning}
\end{figure}

\section{On model architectures of the GAN example in Eqs. \eqref{eq:model_to_data_all} and \eqref{eq:model_to_model_all}}
\label{appsec:GAN_arch}

We next focus on discussing the model architectures of the GAN generator and discriminator employed in Eqs. \eqref{appeq:model_to_data} and \eqref{appeq:model_to_model_v1} (\ie Eqs. \eqref{eq:model_to_data_all} and \eqref{eq:model_to_model_all} of the main manuscript).

Recently, the community begins to exploit integrating ViTs into GANs \cite{jiang2021transgan,lee2021vitgan,zhao2021improved,zhang2021styleswin}.
For example, the ViTGAN \cite{lee2021vitgan}, delivering SOTA generative performance, employs simple modifications to the ViT architecture to construct the generator and the discriminator, but adopts \emph{many} techniques to regularize the ViT-based discriminator for stable training.
Motivated by the modeling flexibility of ViTs, we also employ ViT-based GAN generator and discriminator in the experiments, but similarly, find it challenging to stabilize GAN training with a ViT-based discriminator.
It's worth highlighting that it's possible to design other alternative model architectures for the big learning; we employ what's presented below for a demonstration.

\begin{figure}[H]
	\centering
	\subfloat[GAN Generator]{
		\includegraphics[height=0.5\columnwidth]{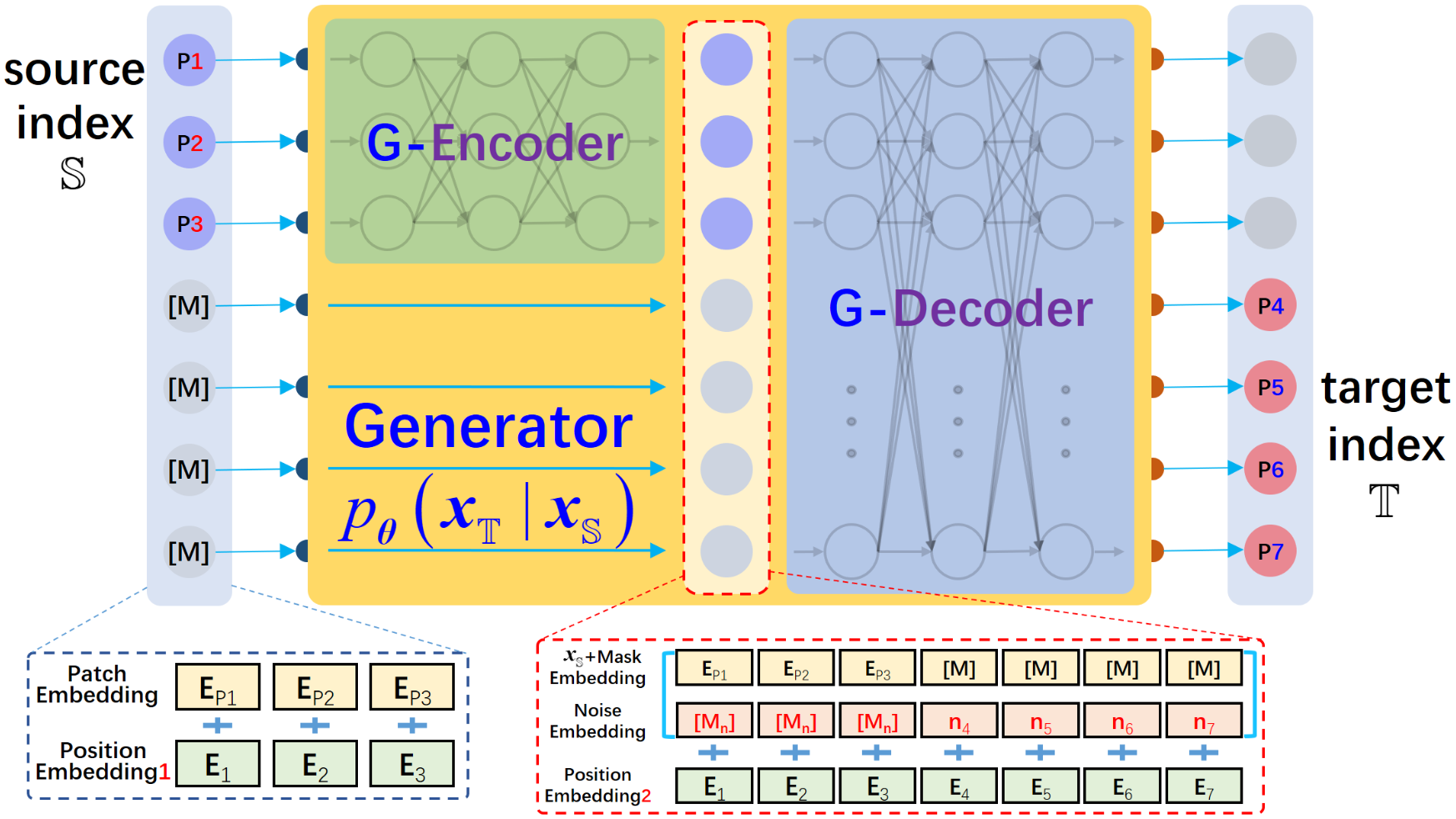}
		\label{appfig:arch_GANgenerator}}
	\qquad
	\subfloat[GAN Discriminator]{
		\includegraphics[height=0.5\columnwidth]{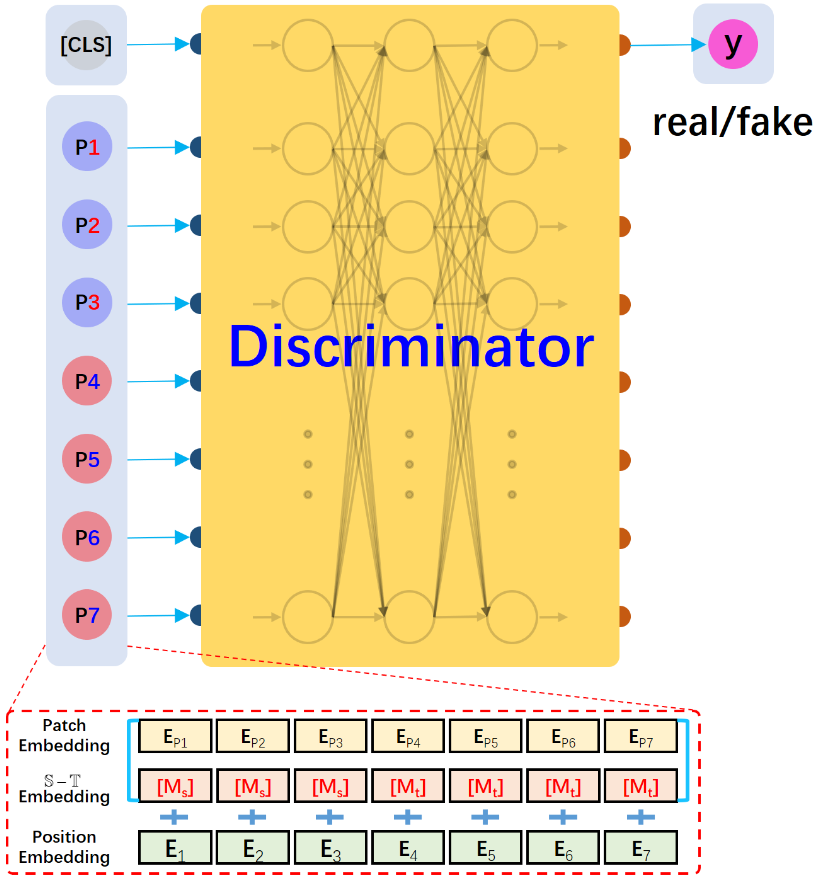}
		\label{appfig:arch_GANdiscriminiator}}
	\caption{Example implementations of the GAN generator and discriminator employed in Eqs. \eqref{appeq:model_to_data} and \eqref{appeq:model_to_model_v1} (\ie Eqs. \eqref{eq:model_to_data_all} and \eqref{eq:model_to_model_all} of the main manuscript). 
	}
	\label{appfig:arch_GAN}
\end{figure}

Fig. \ref{appfig:arch_GAN} demonstrates the employed GAN generator and discriminator, both of which are constructed with Transformers/ViTs to exploit their modeling capabilities and flexibilities.
\begin{itemize}
	\item \textbf{GAN Generator.} 
	Following the MAE \cite{he2021masked}, we design the GAN generator $p_{\thetav}(\xv_{\Tbb} | \xv_{\Sbb})$ with an autoencoder-like architecture, which employs an encoding G-Encoder and a decoding G-Decoder, as shown in Fig. \ref{appfig:arch_GANgenerator}. 
	The G-Encoder encodes the source patches $\xv_{\Sbb}$ (if any) to their latent codes; then, these codes are combined with the mask tokens \texttt{[M]}, patch-wise noise embeddings, and new positional encodings to serve as the input of the G-Decoder; finally, the G-Decoder transforms its input to generate the target patches $\xv_{\Tbb}$. 
	
	$\texttt{[M]}$ tokens are inserted later in a middle layer, because doing this often improves performance and lowers the computational burden \cite{touvron2021going,he2021masked}.
	A noise $\zv$ is mapped with an $8$-layer MLP to produce the patch-wise noise embeddings $\{\nv_1,\cdots,\nv_L\}$. 
	Note we also introduce another toke $\texttt{[M}_{\texttt{n}}{]}$ to indicate no noise embeddings are necessary at the corresponding source locations in $\Sbb$.
	
	\item \textbf{GAN Discriminator.} 
	As shown in Fig. \ref{appfig:arch_GANdiscriminiator}, we also modify the Transformer/ViT architecture to construct the universal GAN discriminator $\sigma(f_{\phiv}(\xv;\Sbb,\Tbb))$ that applies to all $(\Sbb,\Tbb)$ cases. 
	We employ an additional \texttt{CLS} token mimicking the BERT, whose output indicates whether the input patches are realistic or not (more specifically, whether they form a ``real'' data from $q(\xv_{\Sbb\cup\Tbb})$ or a fake one from $p_{\thetav}(\xv_{\Tbb} | \xv_{\Sbb}) q(\xv_{\Sbb})$, by referring to Eq. \eqref{appeq:model_to_data}). 
	The input of the discriminator consists of patch embeddings, positional embeddings, and two new special tokens ($\texttt{[M}_{\texttt{s}}{]}$ and $\texttt{[M}_{\texttt{t}}{]}$) that indicate source or target patches mimicking the sentence tokens in the BERT.
	
\end{itemize}

\section{Big learning versus contrastive learning}
\label{appsec:biglearn_contrastlearn}

Contrastive learning \cite{hadsell2006dimensionality} aims at learning a latent representation space, where the representations of different views of the same image (``positive pairs'') are near each other but those from different images (``negative pairs'') are far away from each other. 

\begin{figure}[H]
	\centering
	\subfloat[Group 1 predicts/retrieves the positively paired sample from the negative samples within a mini-batch. $\Ac$ and $\Bc$ denote the online and target augmentation, respectively.
	]{
		\includegraphics[width=0.8\columnwidth]{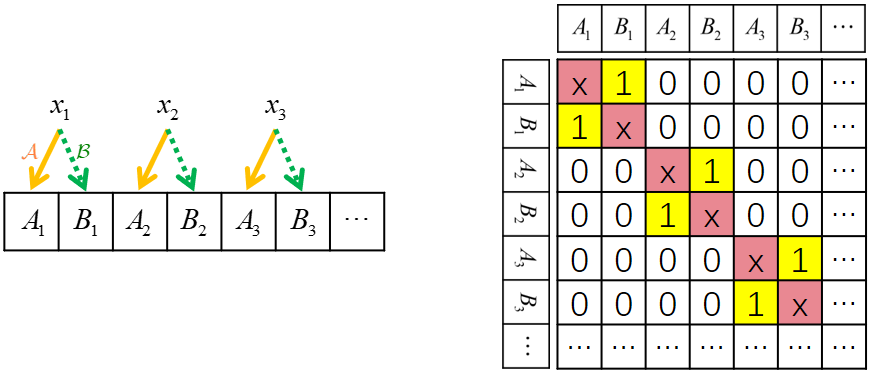}
		\label{appfig:contrastive_retrieve}
	}
	
	\subfloat[Group 2 directly generates/predicts the target/teacher projection conditioned on the student projection. ]{
		\includegraphics[width=0.8\columnwidth]{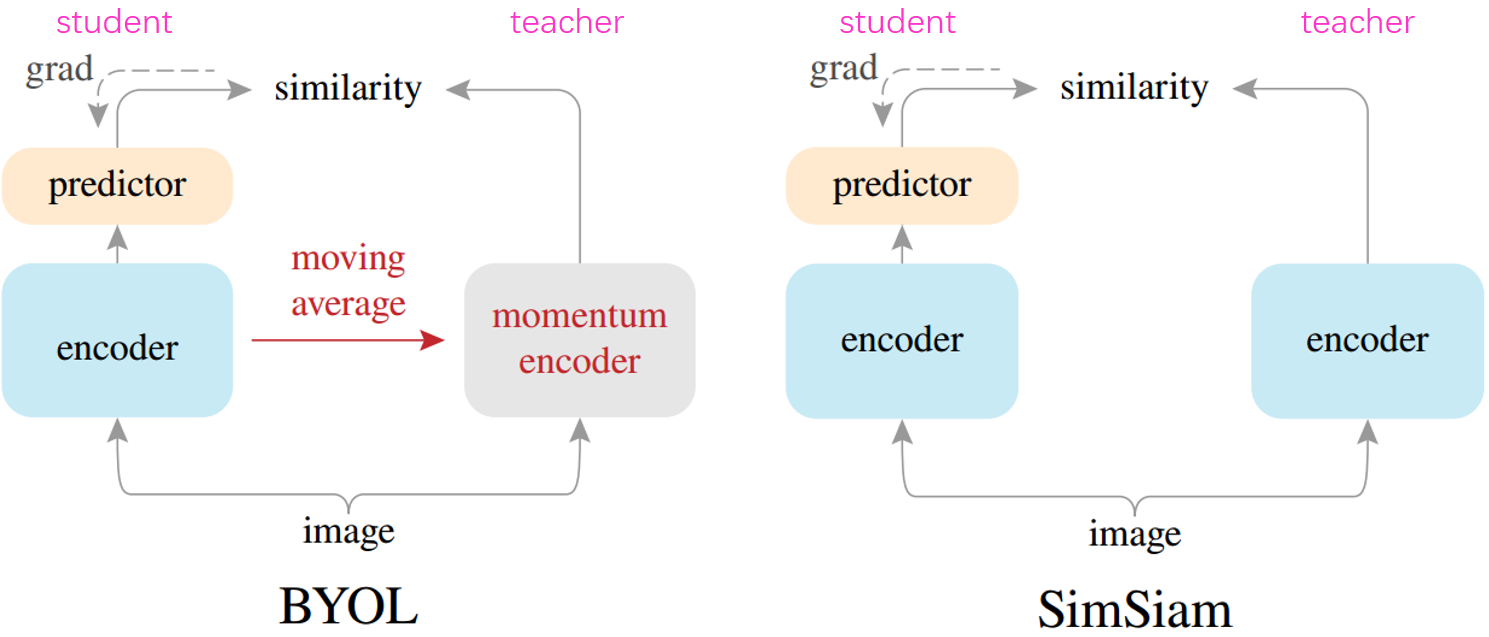}
		\label{appfig:contrastive_generate}
	}
	\caption{Demonstrations of contrastive learning methods. (b) is adapted from Fig. 3 of \cite{chen2021exploring}.
	}
	\label{appfig:contrastive_learning}
\end{figure}

As discussed in the main manuscript, the self-supervised contrastive learning focuses on exploiting domain prior knowledge to learn generally applicable data representations, while the presented big learning is mostly data-driven. From that perspective, they are orthogonal to each other. However, we reveal below that big learning and contrastive learning have a lot in common.
\begin{itemize}
	\item Both of them are based on massive multi-task training, associated with source/target indexes $(\Sbb,\Tbb)$ and online/target augmentation pairs $(\Ac, \Bc)$ (see Fig. \ref{appfig:contrastive_retrieve}), respectively.
	
	\item Both of them share a universal model among massive training tasks.
	
	\item From the information perspective, both of them predict (or retrieve) the $\Tbb/\Bc$-associated information conditioned on the information related to $\Sbb/\Ac$, as detailed below.
\end{itemize}

Existing contrastive learning methods can be roughly grouped into two groups, based on whether the method uses negative pairs (like SimCLR \cite{chen2020simple} and MoCo \cite{chen2021empirical}) or not (like BYOL \cite{grill2020bootstrap} and SimSiam \cite{chen2021exploring}).
\begin{itemize}
	\item \textbf{Group 1.} Contrastive learning methods using negative pairs, like SimCLR and MoCo, can be interpreted as \emph{retrieving} (in the latent representation space) the target positively paired sample $\Bv_{i}$ from the negative samples within the mini-batch, conditioned on the source augmented sample $\Av_{i}$, as illustrated in Fig. \ref{appfig:contrastive_retrieve}.
	
	\item \textbf{Group 2.} Contrastive learning methods not using negative pairs, like BYOL and SimSiam, directly predict/generate (in the latent representation space) the target/teacher projection associated with $\Bv_{i}$, conditioned on the student projection associated with $\Av_{i}$, as demonstrated in Fig. \ref{appfig:contrastive_generate}.

\end{itemize}
Either group of contrastive learning methods retrieves or predicts the $\Bc$-associated information conditioned on the information related to $\Ac$, which is quite similar to the proposed big learning that predicts/generates $\xv_{\Tbb}$ conditioned on $\xv_{\Sbb}$.
Therefore, from the information perspective, both of them predict (or retrieve) a piece of data/prior information conditioned on another piece of data/prior information.

It's interesting to consider combining big learning with contrastive learning to exhaustively exploit the available information from both data and domain-prior perspectives.

\section{Experimental settings used in Sections \ref{sec:unsupervised_exp_Completion} and \ref{sec:unsupervised_exp_abuse} of the main manuscript}
\label{appsec:experiment_settings}

We employ the same model architectures in the previous Section \ref{appsec:GAN_arch} for the experiments on the MNIST and CelebA datasets, with the detailed hyperparameters summarized in Table \ref{apptab:hyperparam_settings}.
Despite the relatively small models used, we find that big learning is capable of delivering potentially all joint/conditional/marginal data capabilities simultaneously.
We adopt the AdamW optimizer \cite{loshchilov2017decoupled} with $\beta=(0.1, 0.999)$ and constant learning rates for both the generator and the discriminator.
Code will be released upon publication.


\begin{table}[htb]
	\centering
	\caption{Hyperparameters used in the experiments.}
		\begin{tabular}{l c c}
			\hline \hline
			Dataset & MNIST & CelebA  \\
			\hline \hline
			Image size & 64 & 120  \\
			Patch size & 8 & 10  \\
			\hline
			G-Encoder depth & 6 & 6  \\
			G-Encoder \#heads & 8 & 8  \\
			G-Encoder dim & 256 & 256  \\
			\hline 
			G-Decoder depth & 6 & 6  \\
			G-Decoder \#heads & 8 & 8  \\
			G-Decoder dim & 512 & 512  \\
			\hline 
			D depth & 6 & 6  \\
			D \#heads & 8 & 8  \\
			D dim & 256 & 256  \\
			\hline 
			GP \cite{mescheder2018training} & real & real  \\
			$\lambda_{\text{GP}}$ & 10 & 10  \\
			Learning rate & $10^{-4}$ & $10^{-4}$  \\
			Batch size & 256 & 128  \\
			\hline 
			Source ratio $\nicefrac{\|\Sbb^1\|}{\|\Lbb\|}$ & Beta(0.5,3) & Beta(0.5,3)  \\
			Target ratio $\nicefrac{\|\Tbb^1\|}{\|\Lbb\backslash\Sbb^1\|}$ & Beta(3,0.5) & Beta(3,0.5)  \\
			Communication source ratio $\nicefrac{\|\Sbb^2\|}{\|\Sbb^1\cup\Tbb^1\|}$ & Beta(0.5,3) & Beta(0.5,3)  \\
 			\hline \hline
		\end{tabular}
	\label{apptab:hyperparam_settings}
\end{table}

%
%
%
%
%
%
%
%
%


%
%
%
%
%
%
%
%
%


Overall, we find it's quite straightforward to implement the MNIST experiments with the standard implementations discussed in Sections \ref{appsec:GAN_derivations} and \ref{appsec:GAN_arch}, without resorting to any ``tricks'' like warm-up or gradient clipping.
However, on the more complicated CelebA experiments, we find it's necessary to employ some, as detailed below.
\begin{itemize}
		
	\item We employ warm-up in the first $10$ epochs for both the GAN generator and discriminator; after that, we use the  constant learning rate given in Table \ref{apptab:hyperparam_settings}.
	
	\item We apply gradient clipping, with the max norm of $5$, to both the generator and discriminator optimizers.
	
	\item Similar to \cite{lee2021vitgan}, we also find it challenging to stabilize GAN training with a ViT-based discriminator. To deal with that, we additionally
	($i$) overlap image patches \cite{lee2021vitgan} with \eg $2$ pixels at the input of the discriminator (different from the non-overlapping image patches used in the vanilla ViT); 
	and 
	($ii$) use a larger hyperparameter $\epsilon=10^{-5}$ in the AdamW optimizer.

\end{itemize}

Other empirical experiences are listed below.
\begin{itemize}

\item We empirically find that the last normalization layers of both the GAN generator and discriminator have a significant influence on the learning stability and final performance. 
Specifically, replacing the last \texttt{LayerNorm} of the G-Decoder of the generator with a \texttt{LeakyReLU} leads to improved generative performance, whereas replacing the last \texttt{LayerNorm} of the discriminator with other normalization/activation layers results in training collapse.

\item Employing an additional convolutional head (like a $3$-layer CNN) to the output of the generator often leads to improved performance and training stability.

\item Instead of only introducing noise embeddings at the first layer of the G-Decoder of the generator, as shown in Fig. \ref{appfig:arch_GANgenerator}, we find it's beneficial to concatenate the same set of noise embeddings layer-wisely into the G-Decoder layers.

\end{itemize}

%
%
%
%
%
%
%
%
%

\section{Big Learning Unifies Classification and Generation}
\label{appsec:biglearn_genclass}

After following \cite{bao2021beit,ramesh2021zero} to vector-quantize an image into discrete tokens $\xv \in \Zbb^{L\times 1}$, the observed random variable $\Xv = (y, \xv)$ with discrete label $y$ now has only one data type. 
Accordingly, one can readily generalize Eq. \eqref{eq:ML_implementation} of the uni-model unsupervised big learning to solve the problem.

Specifically, with a Transformer-based universal model $p_{\thetav}(\Xv_{\bar\Tbb'}|\Xv_{\bar\Sbb'})$ that models the generative process of a target token $\Xv_{\bar\Tbb'}$ given source ones $\Xv_{\bar\Sbb'}$ for \emph{any} $(\bar\Sbb',\bar\Tbb')$ pair, the big learning yields
\beq\label{eq:}
	\max_{\thetav} \Ebb_{q(\Sbb',\Tbb')} \sum_{(\bar\Sbb',\bar\Tbb') \in \Xi^{'}_{\Sbb',\Tbb'}} \Ebb_{q(\Xv_{\bar\Tbb'}|\Xv_{\bar\Sbb'})} \log  p_{\thetav}(\Xv_{\bar\Tbb'}|\Xv_{\bar\Sbb'}),
\eeq
where $q(\Sbb',\Tbb')$ denotes the sampling process of $(\Sbb',\Tbb')$ with random permutations, $\Tbb'=\{t_1, t_2, \cdots\}$, $\Xi^{'}_{\Sbb',\Tbb'} = \{(\Sbb',t_1), (\{\Sbb',t_1\},t_2), (\{\Sbb',t_1, t_2\},t_3), \cdots\}$, often $p_{\thetav}(\Xv_{\bar\Tbb'}|\Xv_{\bar\Sbb'}) =\text{Categorical}(\Xv_{\bar\Tbb'}|\pv_{\thetav}(\Xv_{\bar\Sbb'}))$ is modeled as a categorical distribution with probabilities $\pv_{\thetav}(\Xv_{\bar\Sbb'})$, and $\Xv_{\bar\Tbb}$ always contain one token (either the label $y$ or an $\xv$-token).
Refer also to Table \ref{tab:BigLearn_special_cases} of the main manuscript for other details.

\begin{figure}[H]
	\centering
	\subfloat[Joint Generation]{
		\includegraphics[height=0.39\columnwidth]{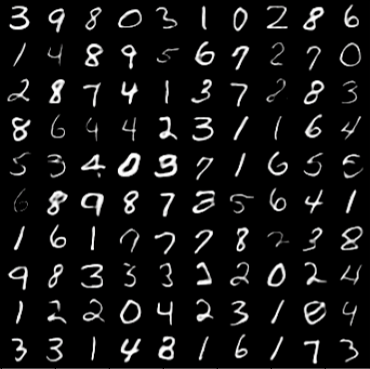}
		\label{fig:}}
	\qquad\quad
	\subfloat[Label-Conditioned Generation]{
		\includegraphics[height=0.39\columnwidth]{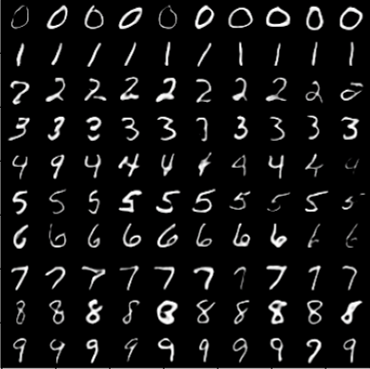}
		\label{fig:}}
	\caption{Demonstration of versatile data capabilities of big learning, retrieved from $p_{\thetav}(\Xv_{\Tbb'}|\Xv_{\Sbb'})$ with specified $(\Sbb', \Tbb')$.
	}
	\label{appfig:biglearning_genclass_all}
\end{figure}

More results complementing Fig. \ref{fig:biglearn_genclass} of the main manuscript are given in Fig. \ref{appfig:biglearning_genclass_all}.

\section{Empirical evaluations on the GLUE benchmark}
\label{appsec:empirical_evaluation}

Concerning the empirical comparisons between existing methods for foundation models and the presented big learning, intuitively, one would consider first using the big learning as the pretraining strategy in place of existing ones, followed by applying the same naïve fine-tuning on downstream tasks, to evaluate the effectiveness of the big learning. 
Unfortunately, we cannot afford the pretraining cost; for example, to pretrain a XLNet-Large takes about $5.5$ days on \textbf{512 TPUs} according to \cite{yang2019xlnet}.
We leave that to the community, as mentioned in the Conclusion.

To demonstrate the advantages of the big learning over existing methods for foundation models, we alternatively consider leveraging it to serve as the less expensive fine-tuning strategy. 
It's worth highlighting that, from another perspective, such experiments also verify the advantages of the big learning in the fields of supervised learning, when compared to existing supervised learning methods.

Specifically, we design experiments based on the Hugging Face transformers library \cite{wolf-etal-2020-transformers}, the GLUE benchmark \cite{wang2018glue}, and the XLNET \cite{yang2019xlnet} that outperforms the BERT on many NLP tasks.
We employ the same pretrained \texttt{xlnet-base-cased} model and continually train it on the downstream RTE/MRPC/SST-2 classification tasks via ($i$) the naive fine-tuning (\ie identical to the original XLNET, termed FT) and ($ii$) the big learning (termed big-learn), respectively. 
In other words, the pretraining phase (\ie the permutation language modeling \cite{yang2019xlnet}, a special case of the big learning) is the same and we compare our big-learn with the naive FT during the finetuning phase.

Because the data of the downstream classification tasks contain both feature $\xv$ and label $y$, we resort to the big learning settings of Section \ref{sec:big_learning} of the main manuscript. Specifically, $\Xv = (\yv, \xv)$ and the universal foundation model $p_{\thetav}(\Xv_{\Tbb'}|\Xv_{\Sbb'})$ has a network architecture similar to the one shown in Fig. \ref{fig:big_learning} of the main manuscript.
Note $p_{\thetav}(\Xv_{\Tbb'}|\Xv_{\Sbb'})$ consists of the pretrained XLNET backbone and a task-specific head that is attached to the output of the \texttt{<CLS>} token; for simplicity, we abuse $\thetav$ to represent all the parameters.  
For a specific $(\Sbb', \Tbb')$ pair, $p_{\thetav}(\Xv_{\Tbb'}|\Xv_{\Sbb'})$ recovers $p_{\thetav}(y|\xv)$, \ie a conventional classifier.

With the above notations, we next formalize the objective for both FT and our big-learn. 
\begin{itemize}
	\item \textbf{FT.} 
	Often a cross-entropy is employed, which is identical to 
	\beq
	\Lc_{\text{FT}}(\thetav) = \Ebb_{q_{\text{downstream}}(\xv,y)} [-\log p_{\thetav}(y|\xv)],
	\eeq
	where $q_{\text{downstream}}(\xv,y)$ represents the training data of the downstream classification task.
	
	\item \textbf{Big-learn.}
	For direct comparisons, we formalize the big-learn objective as 
	\beq\label{appeq:big_learn_1}
		\Lc_{\text{big-learn}}(\thetav) = \Lc_{\text{FT}}(\thetav) + \beta_{\text{BigLearn}} \Lc(\thetav),
	\eeq
	where $\beta_{\text{BigLearn}}$ is a hyperparameter and 
	\beq\label{appeq:big_learn_0}
		\Lc(\thetav) = \Ebb_{q(\Sbb', \Tbb')} \Ebb_{q_{\text{downstream}}(\Xv)} [-\log p_{\thetav}(\Xv_{\Tbb'}|\Xv_{\Sbb'})],
	\eeq
	with $q(\Sbb', \Tbb')$ denoting the sampling process of $(\Sbb', \Tbb')$.
	We simply reuse the same sampling process in Table \ref{apptab:hyperparam_settings}. 
\end{itemize}

Note Eq. \eqref{appeq:big_learn_0} is equivalent to minimizing $\Ebb_{q(\Sbb', \Tbb')}\KL[q_{\text{downstream}}(\Xv_{\Tbb'}|\Xv_{\Sbb'})||p_{\thetav}(\Xv_{\Tbb'}|\Xv_{\Sbb'})]$ by referring to Eq. \eqref{eq:cross_entropy_loss} of the main manuscript.

\begin{table}[H]
	\centering
	\caption{Tested hyperparameters when comparing FT with big-learn on the GLUE benchmark.}
		\begin{tabular}{l c c c c}
			\hline \hline
			{Task}$\backslash${Hyperparameter} & Learning Rate & \#Epochs & WarmUp Steps & $\beta_{\text{BigLearn}}$
			\\ \hline 
			RTE & [2e-5, 4e-5, 6e-5] & [3, 4, 7, 10, 15] & [0, 120] & [0., 0.2, 0.4, 0.6, 0.8] 
			\\
			MRPC & [2e-5, 4e-5, 6e-5] & [3, 4, 7, 10, 15] & [0, 120] & [0., 0.2, 0.4, 0.6, 0.8] 
			\\
			SST-2 & [2e-5, 4e-5, 6e-5] & [2, 3, 4] & [0, 1200] & [0., 0.2, 0.4] 
			\\ \hline \hline
		\end{tabular}
	\label{apptab:glue_hyperpara}
\end{table}

We extensively compare FT with big-learn on the downstream RTE/MRPC/SST-2 classification tasks, by evaluating the accuracy and/or F1 score on the Dev set across the combinations of the tested hyperparameters shown in Table \ref{apptab:glue_hyperpara}. 
The hyperparameters are chosen following \cite{devlin2018bert,yang2019xlnet}.

\begin{figure}[H]
	\centering
	\subfloat[RTE]{
		\includegraphics[height=0.33\columnwidth]{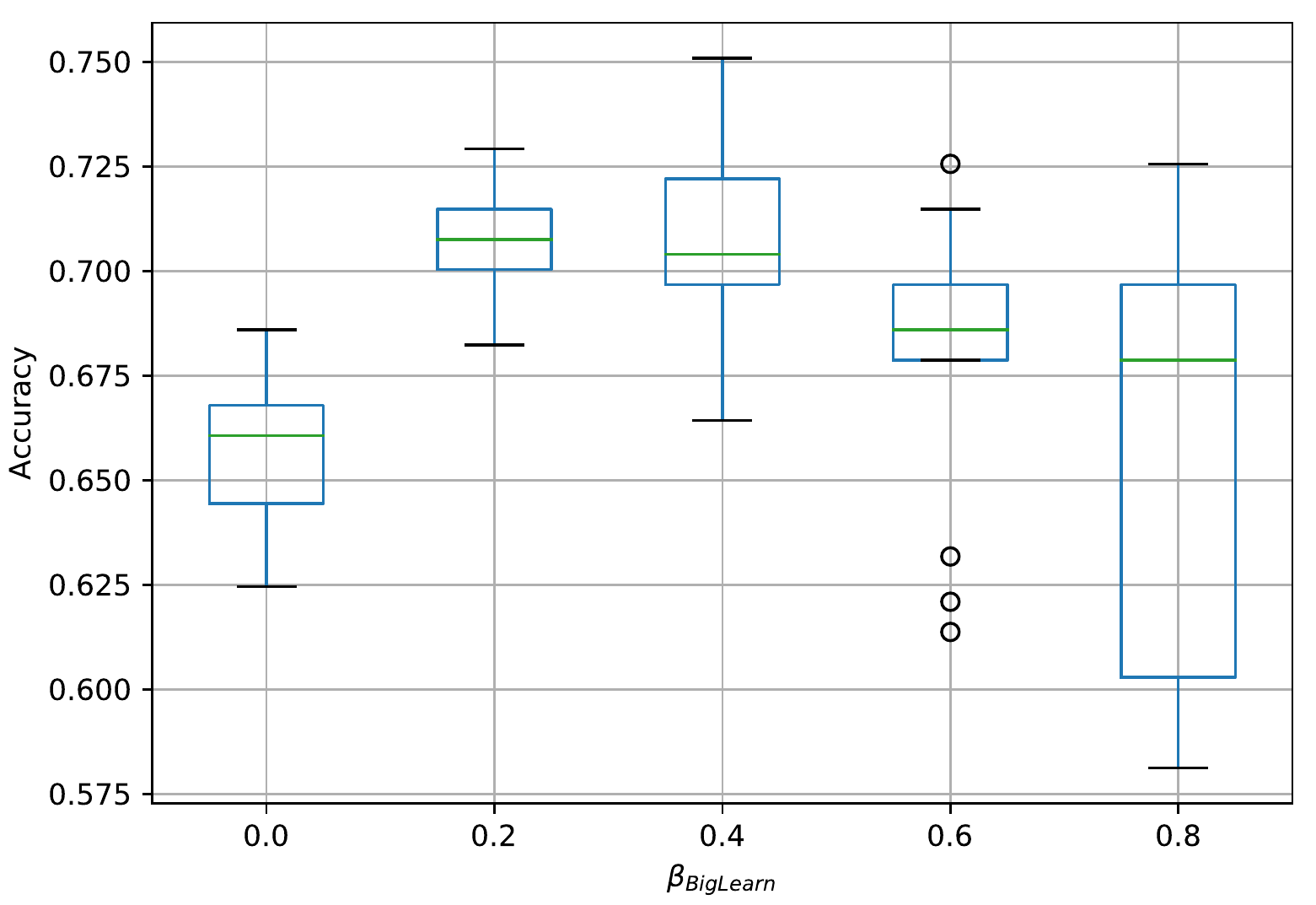}
		\label{fig:}}
	\subfloat[MRPC]{
		\includegraphics[height=0.33\columnwidth]{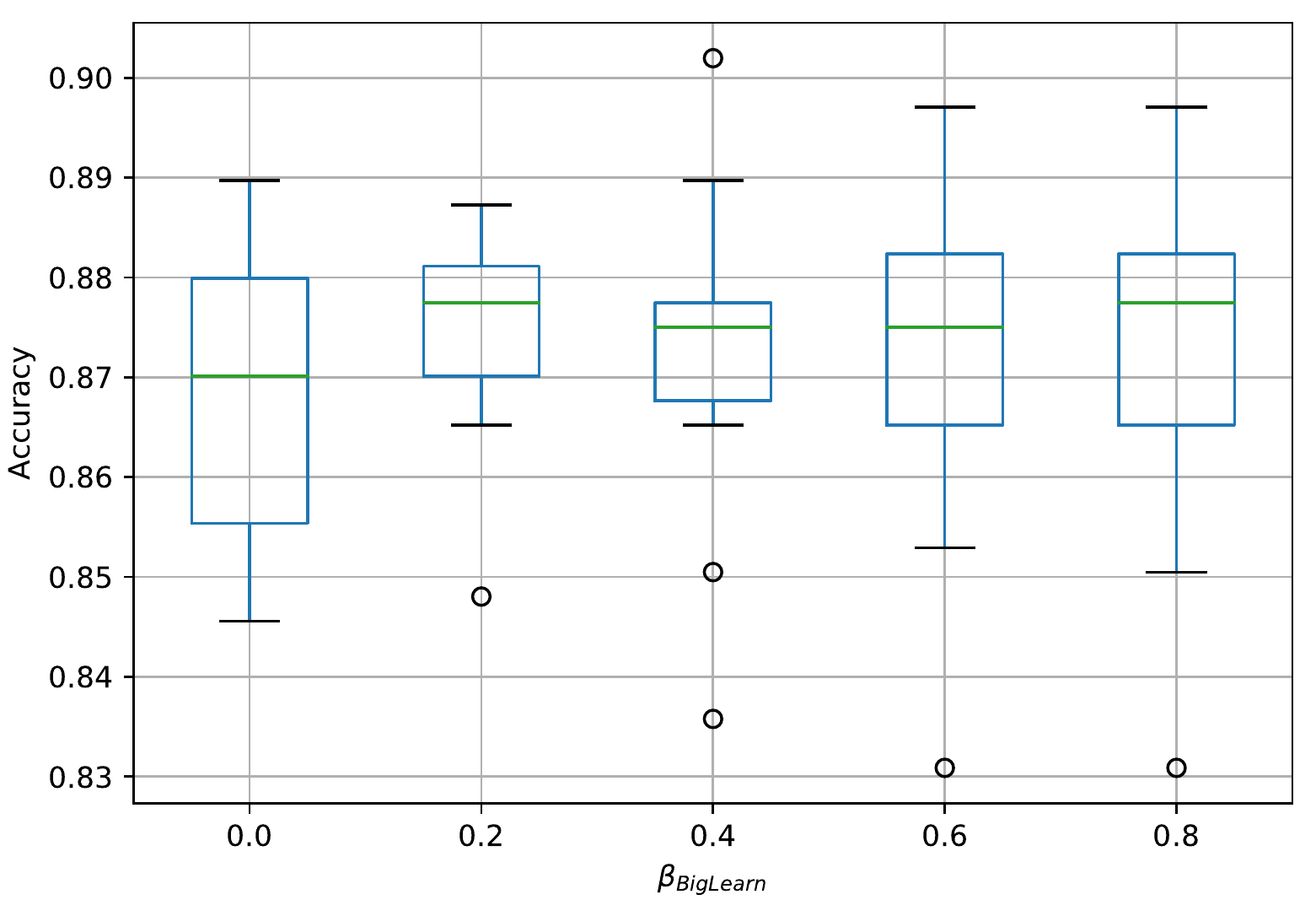}
		\label{fig:}}
	\qquad
	\subfloat[SST-2]{
	\includegraphics[height=0.33\columnwidth]{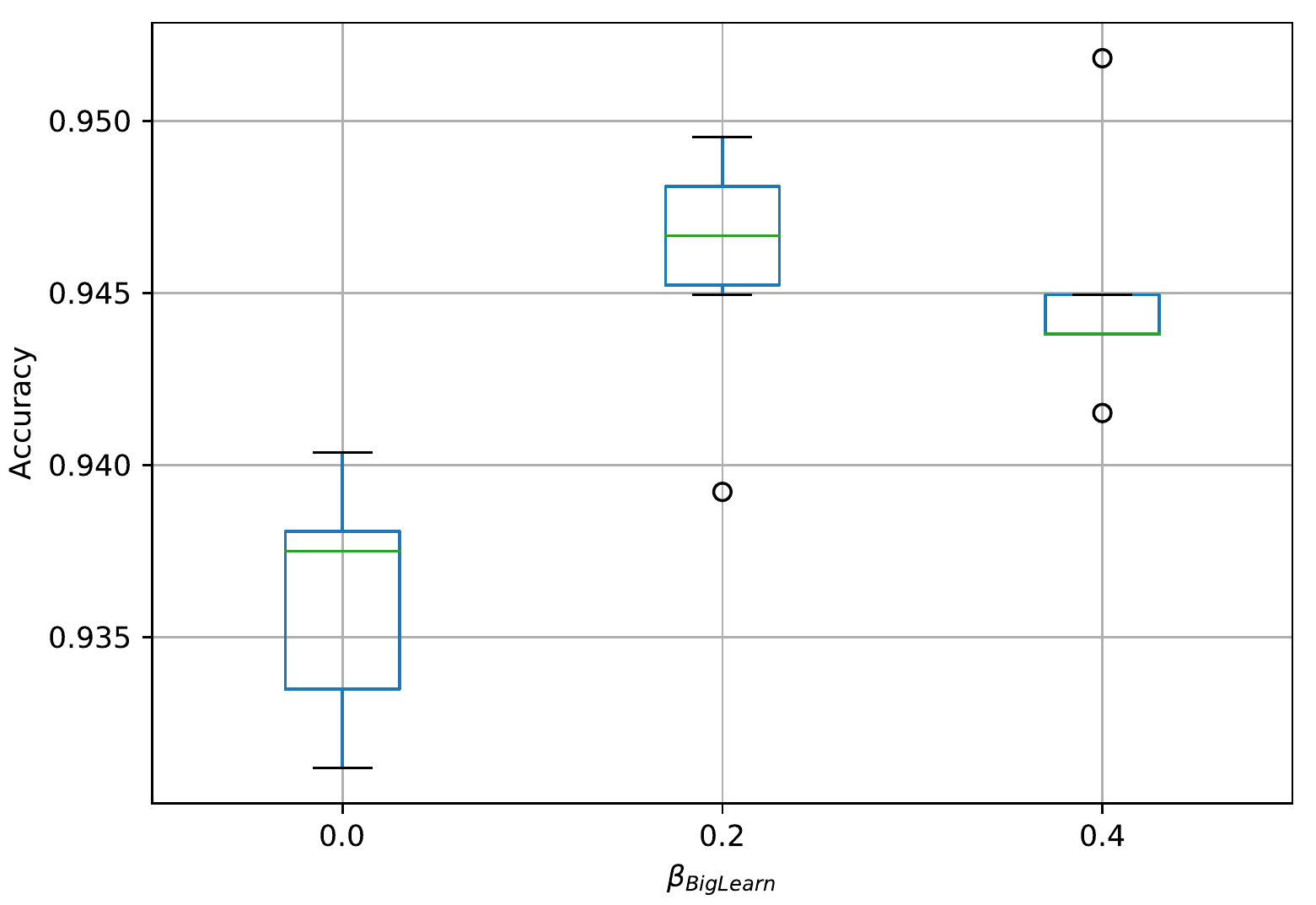}
	\label{fig:}}
	\caption{Boxplots of the Dev-set accuracies from FT and our big-learn. Note big-learn with $\beta_{\text{BigLearn}}=0$ is identical to FT (see Eq. \eqref{appeq:big_learn_1}).
		It's clear that big-learn consistently outperforms FT on all three tasks.
	}
	\label{appfig:boxplot_glue}
\end{figure}

The best/median metrics are summarized in Table \ref{apptab:glue_ACC} and Fig. \ref{appfig:boxplot_glue} shows the corresponding boxplots; it's clear that our big-learn consistently outperforms FT. Accordingly, the big learning can serve as a superior fine-tuning strategy.
It's worth highlighting we did not carefully tune our big-learn; therefore, it's likely that its performance could be further improved by \eg tuning the sampling process $q(\Sbb', \Tbb')$.

We'd like to emphasize that the big learning can reduce the pretrain-finetuning gap because 
\begin{itemize}
	\item it can act as the pretraining and finetuning objectives, simultaneously;
	\item one can even rely on the big learning to completely merge the pretraining and finetuning phases, leading to a zero gap.
\end{itemize}

Motivated by the performance boost from the BERT to the XLNET and our discussions ``on the generalization of model parameters and latent features'' of Section 3.2 of the main manuscript, we posit that the big learning can serve as better pretraining and finetuning strategies than existing methods, leading to a universal machine learning paradigm. 
We leave the corresponding verification as future research.

\section{Additional experimental results}
\label{appsec:add_exp_results}

More experimental results, complementing the limited demonstrations of the main manuscript, are given below. Please refer to the captions for details.

\begin{figure}[H]
	\centering
	\includegraphics[width=1.2\columnwidth]{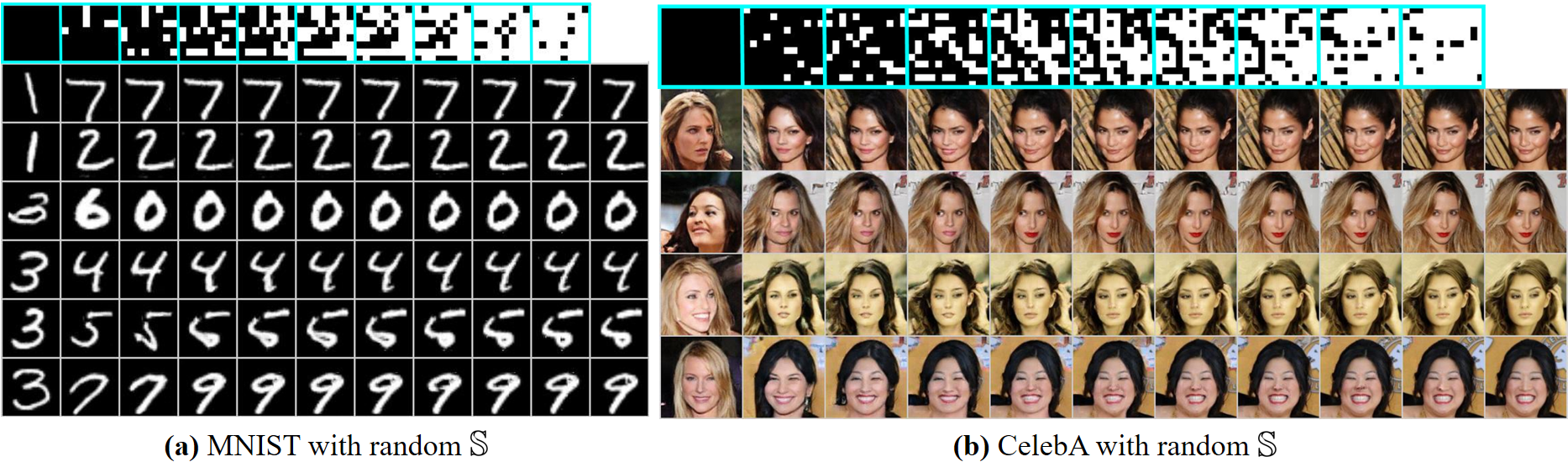}
	\includegraphics[width=1.2\columnwidth]{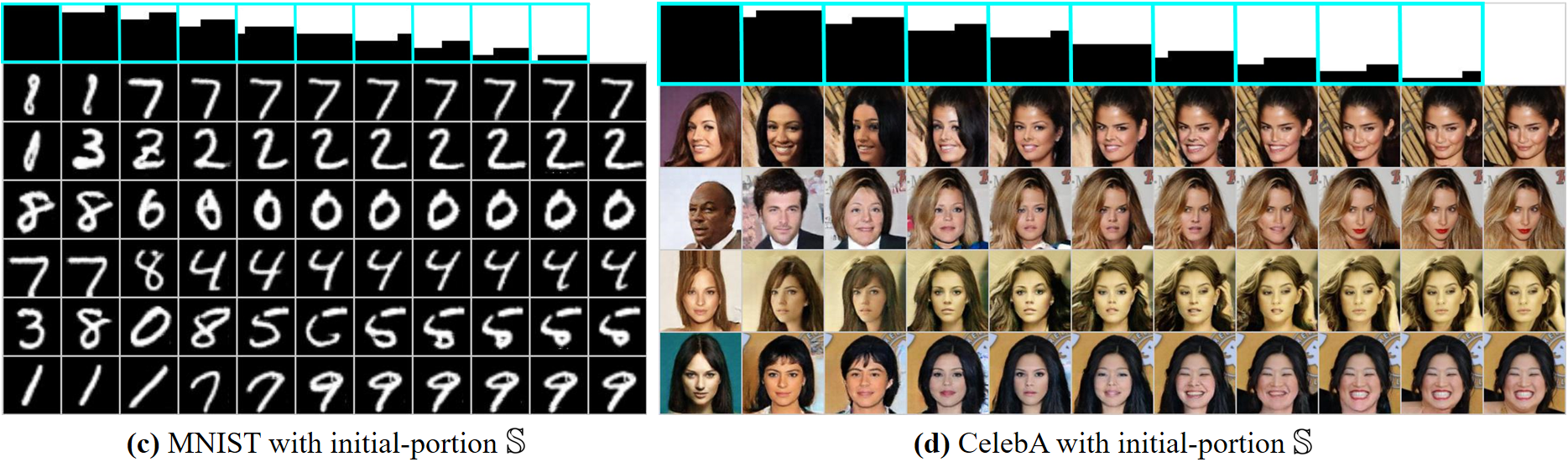}
	\caption{Demonstrating the generation/completion capabilities of big learning when gradually increasing the ratio of $\Sbb$ from $0$ (joint generation) to $0.9$, from left to right.
		Shown in the light-blue boxes of the first row are the masks of $\xv_{\Sbb}$ applied in each column; white/black indicates $\Sbb/\Tbb$. 
		The right-most column shows ground-truth $\xv$ shared in each row.
		Note each row also employs the same noise.
		It's clear that the generations become increasingly similar/dissimilar to the ground-truth $\xv$ as the ratio of $\Sbb$ increases/decreases, as expected. 
		See the category, style, and thickness of the MNIST generations as the ratio of $\Sbb$ decreases, as well as the identity, expression, hairstyle, and gender of the CelebA generations.
		Big learning produces realistic and diverse generations/completions in all situations.
	}
	\label{fig:diff_Sratio}
\end{figure}

\begin{figure}[H]
	\centering
	\includegraphics[width=1\columnwidth]{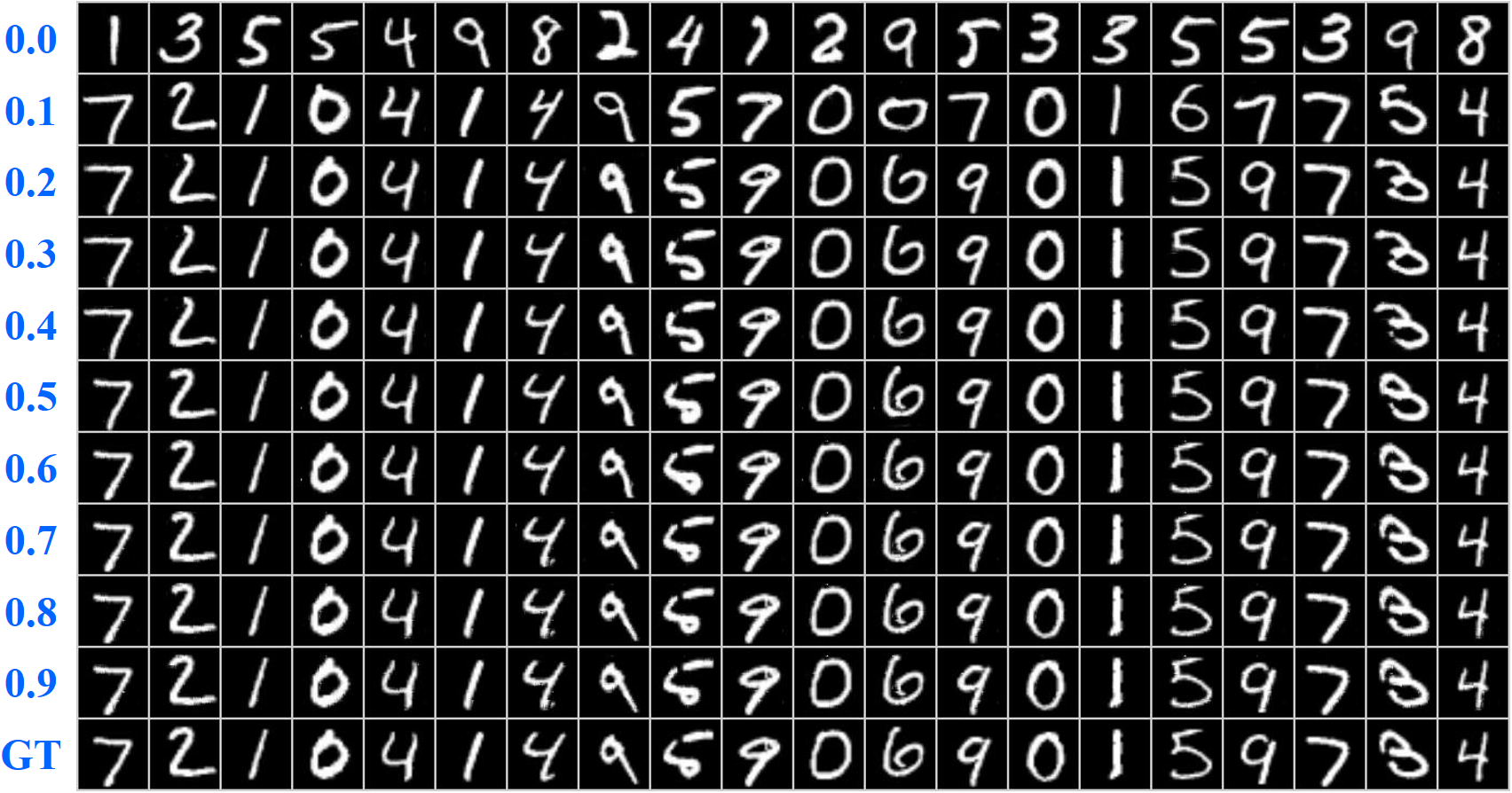}
	\caption{More MNIST generations/completions from big learning when gradually increasing the ratio of $\Sbb$ from $0.0$ to $0.9$.
	}
	\label{fig:}
\end{figure}

\begin{figure}[H]
	\centering
	\includegraphics[width=1\columnwidth]{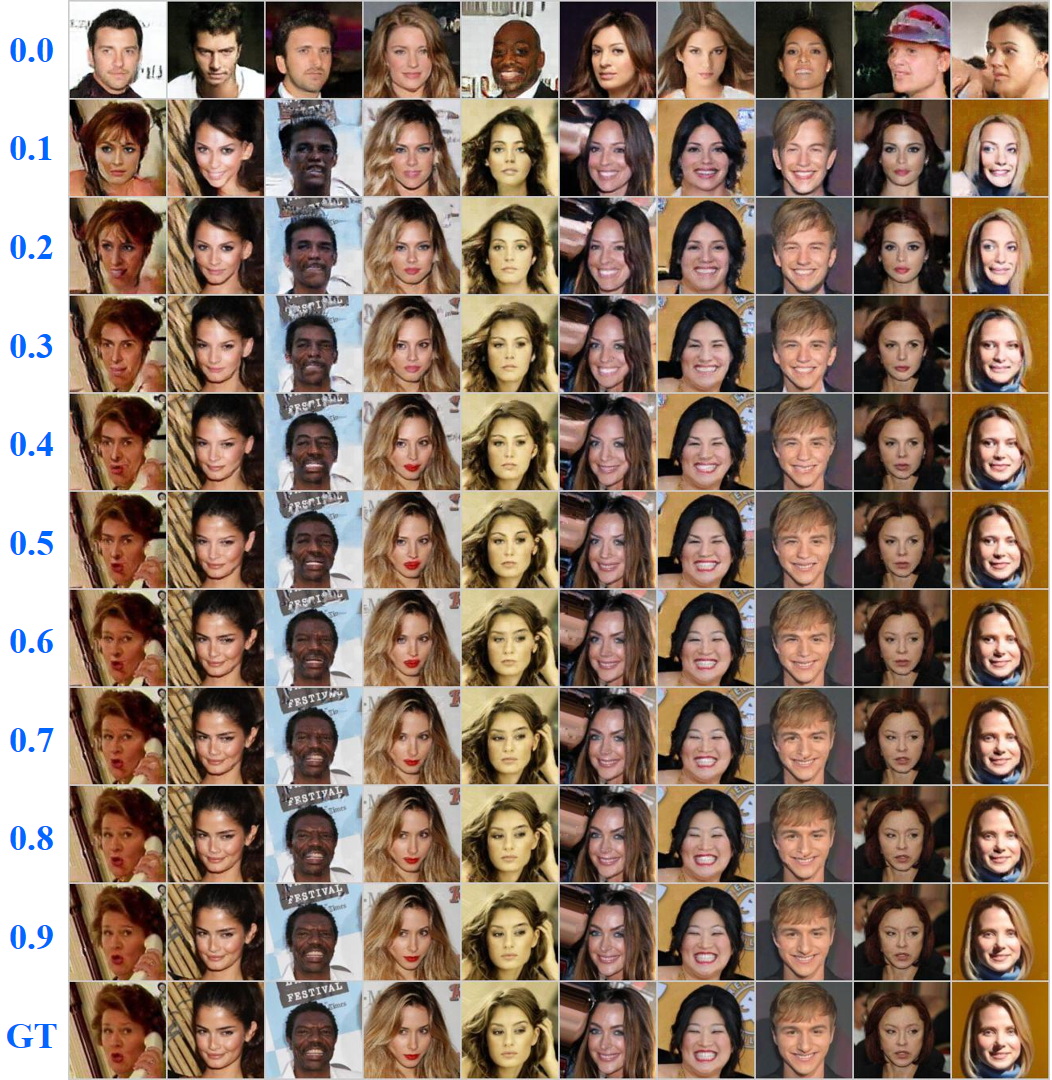}
	\caption{More CelebA generations/completions from big learning when gradually increasing the ratio of $\Sbb$ from $0.0$ to $0.9$.
	}
	\label{fig:}
\end{figure}

\begin{figure}[H]
	\centering
	\includegraphics[width=1.2\columnwidth]{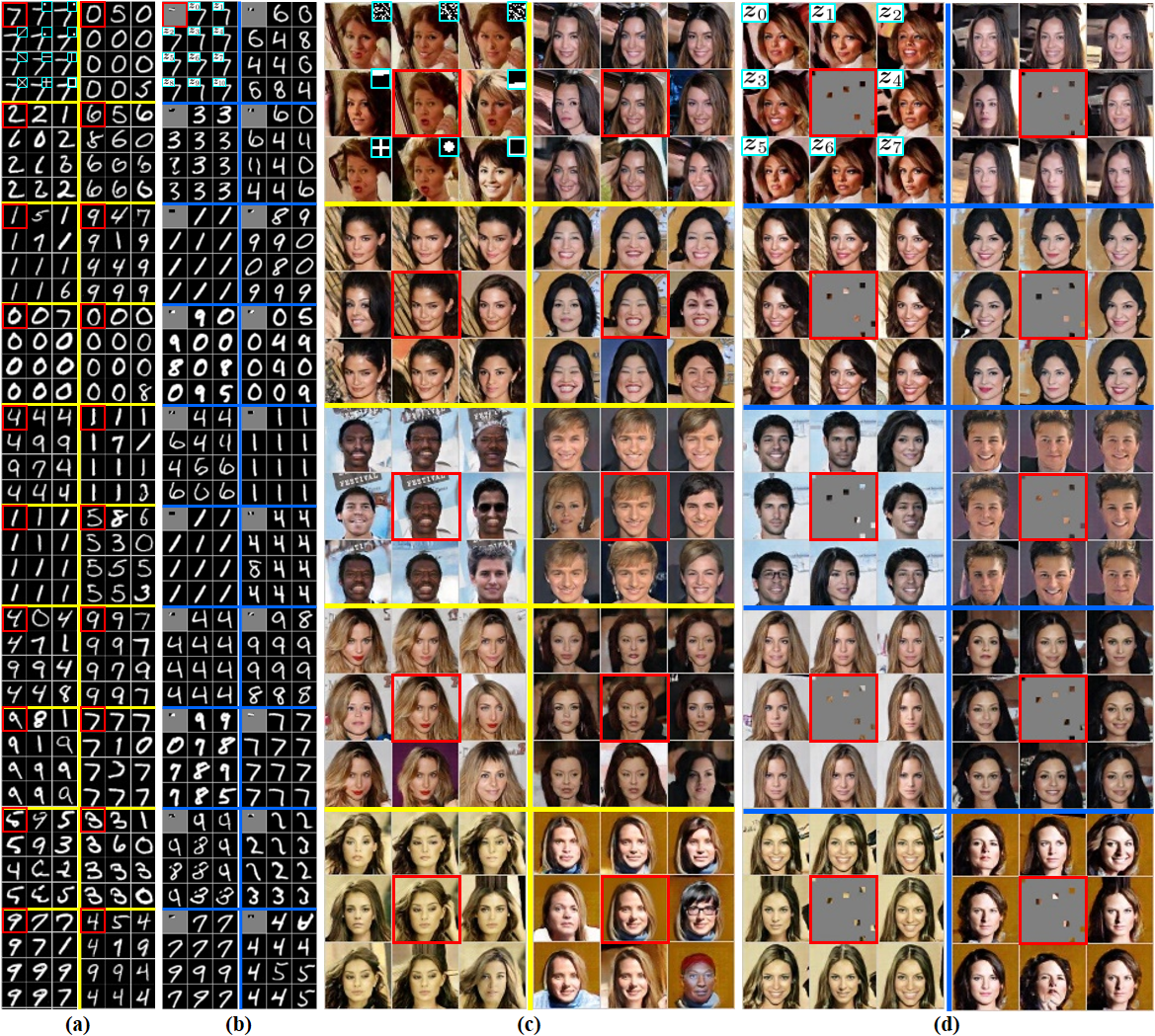}
	\caption{The diverse generations/completions of big learning with (a)(c) various $\Sbb$ settings and (b)(d) different noises. 
		Shown in red boxes are either the ground-truth images $\xv$ or the source $\xv_{\Sbb}$. 
		Big learning delivers diverse realistic generations \wrt different $\Sbb$/noise settings.
	}
	\label{fig:}
\end{figure}

\begin{figure}[H]
	\centering
	\includegraphics[width=0.9\columnwidth]{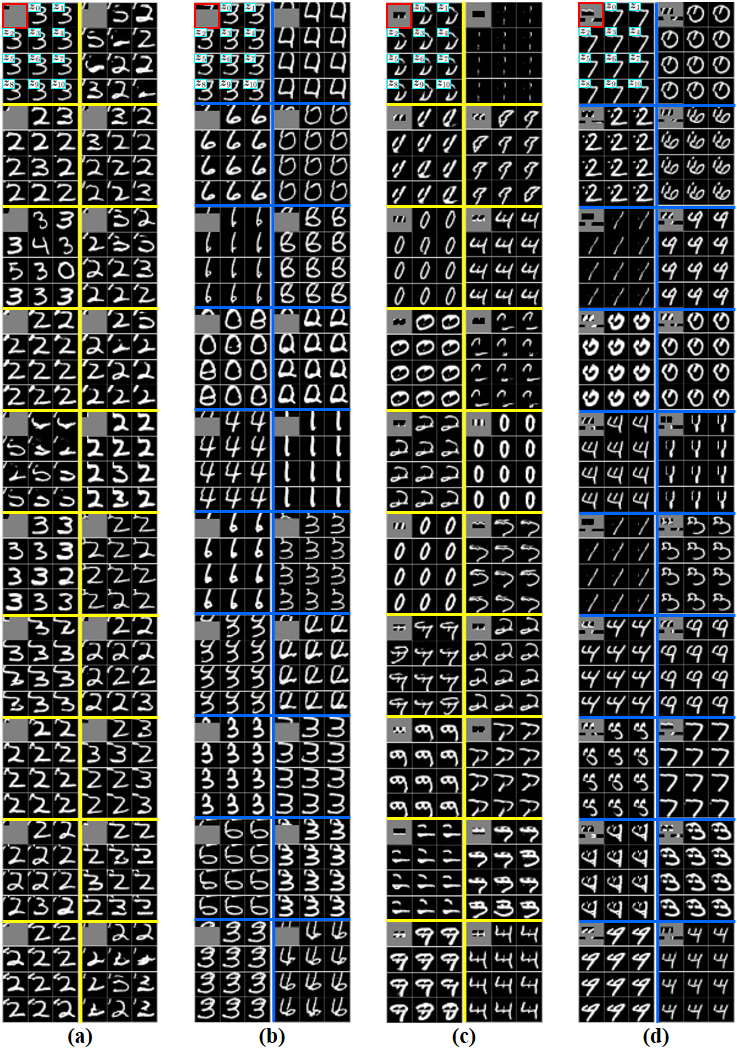}
	\caption{The strong generalization capability of big learning \wrt anomalous testing cases out of the training domain. 
		Big learning generalizes well on $\xv_{\Sbb}$s that are constructed with  
		(a) random center patches replaced in the upper-left corner,
		(b) random center patches replaced in the upper part,
		(c) random center patches duplicated and replaced in the center,
		and
		(d) random patches and more complicated manipulations (including duplication, relocation, and mix-up).
	}
	\label{fig:}
\end{figure}

\begin{figure}[H]
	\centering
	\includegraphics[width=1.2\columnwidth]{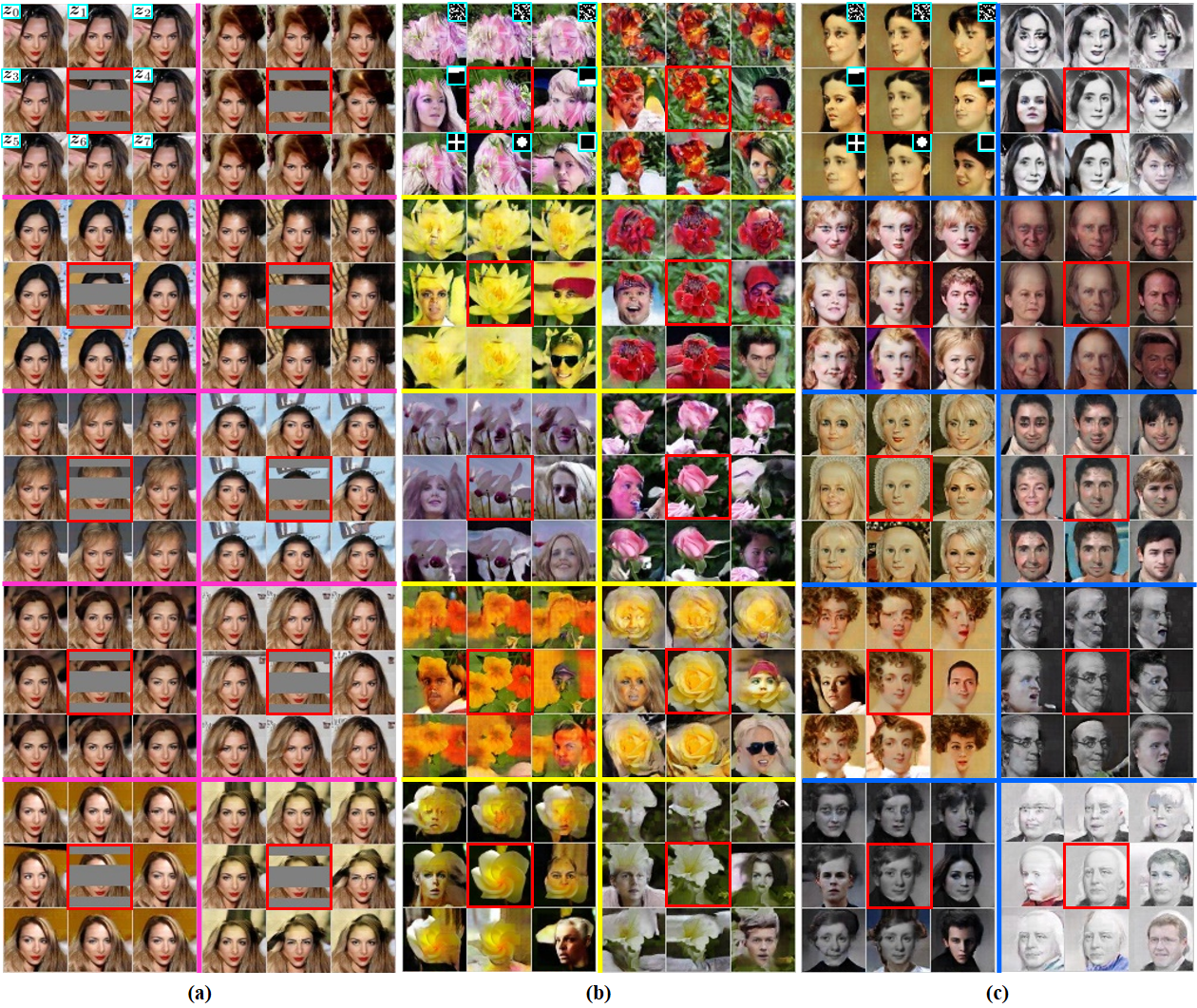}
	\caption{The strong generalization capability of big learning \wrt anomalous/unseen testing cases out of the training domain, on (a) CelebA, (b) Flowers, and (c) MetFaces. 
		Big learning generalizes well on $\xv_{\Sbb}$ constructed by  
		(a) mixing-up patches from different CelebA images,
		(b) sampling out-of-domain image patches from the Flowers dataset,
		and 
		(c) sampling out-of-domain image patches from the MetFaces dataset.
	}
	\label{fig:}
\end{figure}

\begin{figure}[H]
	\centering
	\includegraphics[width=1\columnwidth]{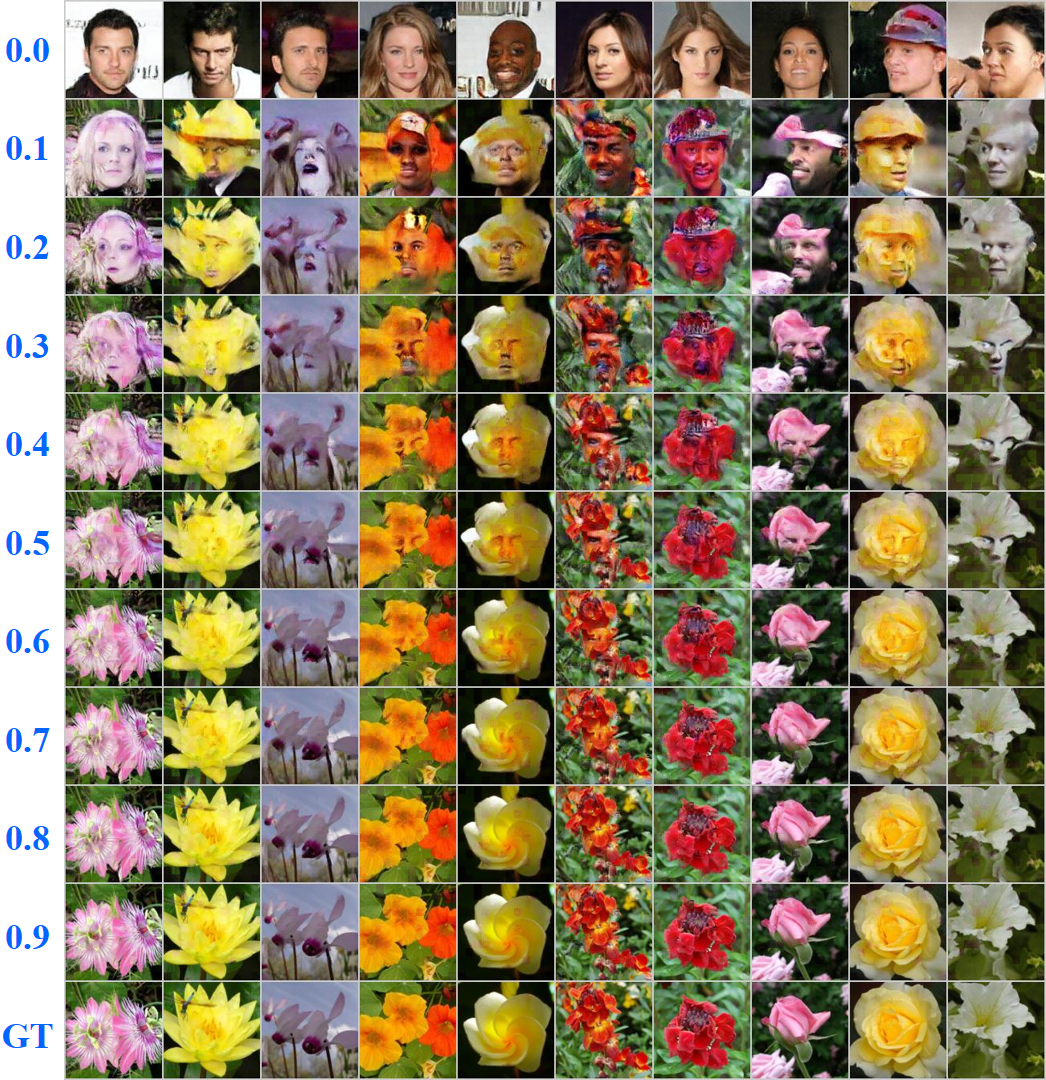}
	\caption{Out-of-domain generations/completions from big learning on the Flowers, when gradually increasing the ratio of $\Sbb$ from $0.0$ to $0.9$.
		The tested model is big-learned on the CelebA.
	}
	\label{fig:}
\end{figure}

\begin{figure}[H]
	\centering
	\includegraphics[width=1\columnwidth]{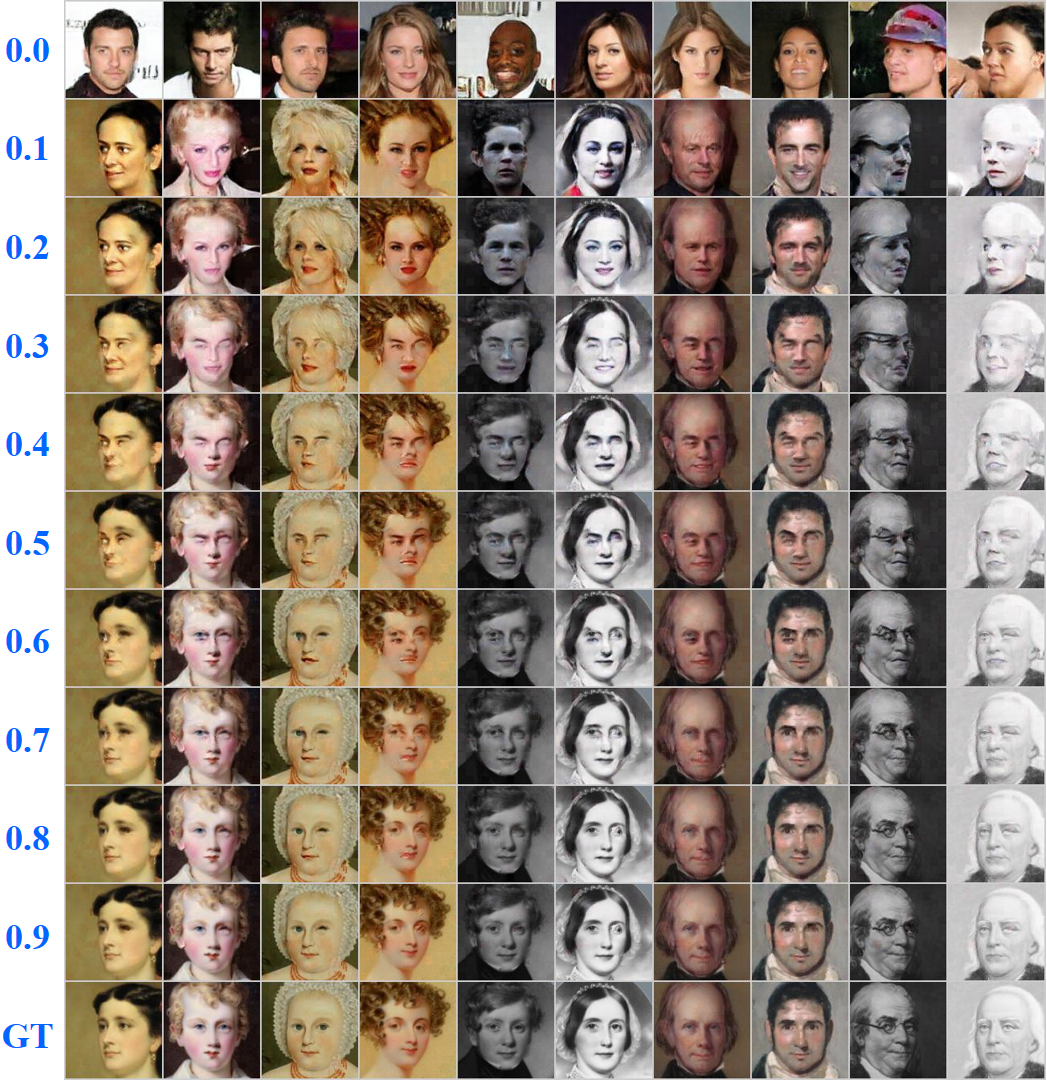}
	\caption{Out-of-domain generations/completions from big learning on the MetFaces, when gradually increasing the ratio of $\Sbb$ from $0.0$ to $0.9$.
		The tested model is big-learned on the CelebA.
	}
	\label{fig:}
\end{figure}

\end{document}